\documentclass[10pt,logo,copyright]{nvidiatechreport}
\usepackage[authoryear,round]{natbib}

\usepackage[utf8]{inputenc} 
\usepackage[T1]{fontenc}    

\usepackage{parskip}        
\usepackage{url}            
\usepackage{booktabs}       
\usepackage{amsfonts}       
\usepackage{nicefrac}       
\usepackage{microtype}      
\usepackage{xcolor}         
\usepackage[dvipsnames]{xcolor} 
\usepackage{graphicx}
\usepackage{animate}        
\usepackage{subcaption}
\usepackage{tabularx}
\usepackage{makecell}
\usepackage{adjustbox}
\usepackage{setspace}
\newcolumntype{M}[1]{>{\centering\arraybackslash}m{#1}}
\usepackage{float}
\usepackage{tikz}
\usetikzlibrary{positioning,shapes,arrows}
\usepackage{amsmath,amsfonts,bm, bbm,leftindex}
\usepackage{multirow}
\usepackage{comment}
\usepackage{gensymb}
\usepackage{lipsum}
\usetikzlibrary{arrows.meta, positioning, fit}
\usepackage[para]{threeparttable}
\usepackage{tikz}
\usetikzlibrary{tikzmark}
\usepackage{hyperref}
\usepackage[most]{tcolorbox}
\usepackage{fancyvrb}
\usepackage{fvextra}
\usepackage{dashrule}
\usepackage{wrapfig}

\def\eqref#1{equation~\ref{#1}}

\def\1{\bm{1}}

\DeclareMathAlphabet{\mathsfit}{\encodingdefault}{\sfdefault}{m}{sl}
\SetMathAlphabet{\mathsfit}{bold}{\encodingdefault}{\sfdefault}{bx}{n}

\makeatletter
\let\save@mathaccent\mathaccent
\newcommand*\if@single[3]{%
  \setbox0\hbox{${\mathaccent"0362{#1}}^H$}%
  \setbox2\hbox{${\mathaccent"0362{\kern0pt#1}}^H$}%
  \ifdim\ht0=\ht2 #3\else #2\fi
  }
\newcommand*\rel@kern[1]{\kern#1\dimexpr\macc@kerna}
\newcommand*\widebar[1]{\@ifnextchar^{{\wide@bar{#1}{0}}}{\wide@bar{#1}{1}}}
\newcommand*\wide@bar[2]{\if@single{#1}{\wide@bar@{#1}{#2}{1}}{\wide@bar@{#1}{#2}{2}}}
\newcommand*\wide@bar@[3]{%
  \begingroup
  \def\mathaccent##1##2{%
    \let\mathaccent\save@mathaccent
    \if#32 \let\macc@nucleus\first@char \fi
    \setbox\z@\hbox{$\macc@style{\macc@nucleus}_{}$}%
    \setbox\tw@\hbox{$\macc@style{\macc@nucleus}{}_{}$}%
    \dimen@\wd\tw@
    \advance\dimen@-\wd\z@
    \divide\dimen@ 3
    \@tempdima\wd\tw@
    \advance\@tempdima-\scriptspace
    \divide\@tempdima 10
    \advance\dimen@-\@tempdima
    \ifdim\dimen@>\z@ \dimen@0pt\fi
    \rel@kern{0.6}\kern-\dimen@
    \if#31
      \overline{\rel@kern{-0.6}\kern\dimen@\macc@nucleus\rel@kern{0.4}\kern\dimen@}%
      \advance\dimen@0.4\dimexpr\macc@kerna
      \let\final@kern#2%
      \ifdim\dimen@<\z@ \let\final@kern1\fi
      \if\final@kern1 \kern-\dimen@\fi
    \else
      \overline{\rel@kern{-0.6}\kern\dimen@#1}%
    \fi
  }%
  \macc@depth\@ne
  \let\math@bgroup\@empty \let\math@egroup\macc@set@skewchar
  \mathsurround\z@ \frozen@everymath{\mathgroup\macc@group\relax}%
  \macc@set@skewchar\relax
  \let\mathaccentV\macc@nested@a
  \if#31
    \macc@nested@a\relax111{#1}%
  \else
    \def\gobble@till@marker##1\endmarker{}%
    \futurelet\first@char\gobble@till@marker#1\endmarker
    \ifcat\noexpand\first@char A\else
      \def\first@char{}%
    \fi
    \macc@nested@a\relax111{\first@char}%
  \fi
  \endgroup
}
\makeatother

\usepackage{url}
\usepackage{booktabs}
\usepackage{graphicx}
\usepackage{multirow}
\usepackage{xcolor}
\usepackage{amsmath}
\usepackage{amssymb}
\usepackage{algorithm}
\usepackage{algpseudocode}
\usepackage[normalem]{ulem}
\usepackage{tabularx}
\usepackage{array}
\usepackage{booktabs}
\usepackage{pifont}
\usepackage{makecell}
\usepackage{colortbl}
\usepackage{svg}
\usepackage{etoc}
\usepackage{wrapfig}

\usepackage{pifont}

\usepackage[nameinlink]{cleveref}
\crefname{equation}{Eq.}{Eqs.}
\crefname{figure}{Fig.}{Figs.}
\crefname{section}{Sec.}{Sec.}
\crefname{appendix}{App.}{App.}
\crefname{table}{Tab.}{Tabs.}
\crefname{algorithm}{Algo}{Algo}
\crefname{thm}{Thm}{Thm}
\Crefname{thm}{Thm}{Thm}
\crefname{prop}{Prop}{Prop}
\makeatletter
\AddToHook{cmd/appendix/before}{\def\cref@section@alias{appendix}}
\makeatother

\usepackage{gradient-text}
\definecolor{darkgreen}{RGB}{0, 154, 85}
\definecolor{nvgreen}{RGB}{136,204,2}

\newcommand{\VeriFine}{%
    \textsc{\textbf{\gradientRGB{VeriFine}{1,111,214}{121,199,5}}}%
}

\newif\ifshowtodos
\showtodostrue     
\ifshowtodos
  \newcommand{\todo}[1]{\textcolor{red}{[\textit{TODO: #1}]}\xspace}
\else
  \newcommand{\todo}[1]{}
\fi

\definecolor{darkred}{rgb}{0.7, 0.0, 0.0}

\newcommand{\crefnames}[3]{%
  \@for\next:=#1\do{%
    \expandafter\crefname\expandafter{\next}{#2}{#3}%
  }%
}

\usepackage[most]{tcolorbox}
\tcbuselibrary{listings,skins,breakable}

\newtcolorbox{promptbox}[2][]{enhanced,
  breakable,
  colback=gray!5,
  colframe=gray!60,
  fonttitle=\bfseries,
  coltitle=black,
  title={Prompt~\thetcbcounter: #2},
  label={#1},
  boxrule=0.5pt,
  arc=2pt,
  left=6pt,
  right=6pt,
  top=6pt,
  bottom=6pt,
}

\title{\raisebox{-.16\height}{\includegraphics[width=0.07\textwidth]{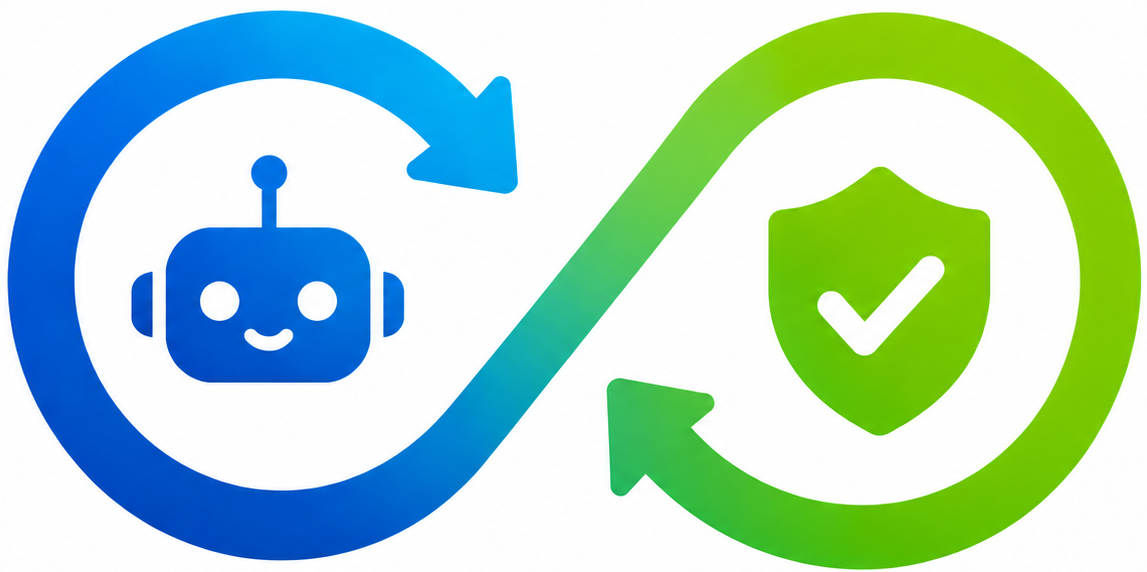}} \hspace{-0.2em} \VeriFine: Scaling Verification for Self-Improvement in Embodied Reasoning}

\author[1,2,*]{Zewei Zhou}
\author[1]{Rachel Luo}
\author[1]{Yulong Cao}
\author[1]{Chaowei Xiao}
\author[1,3]{Chensheng Peng}
\author[1]{Boyi Li}
\author[1]{Thomas Tian}
\author[1]{Zheng Lian}
\author[1]{Yan Wang}
\author[1]{Jiaqi Ma}
\author[1]{Boris Ivanovic}
\author[1,4]{Marco Pavone}
\author[1,$\dagger$]{Wenhao Ding}

\affil[1]{%
  NVIDIA\quad
  \textsuperscript{2}UCLA\quad
  \textsuperscript{3}UC Berkeley\quad
  \textsuperscript{4}Stanford University\quad\quad
  \textsuperscript{*}Work done during an internship at NVIDIA\quad
  \textsuperscript{$\dagger$}Corresponding author%
}

\begin{abstract}
Self-improving policies continually expose new failure patterns, changing what their judges must be able to verify. However, current fixed judges constrain both optimization feedback and the discovery of useful training examples, limiting further self-improvement. This challenge is even more acute in embodied reasoning, where reliable evaluation must account for spatial grounding, causal reasoning, and safety-aware decision-making. We introduce \VeriFine, an agent harness framework that scales verification through the co-evolution of the policy, training curriculum, and judge. The \textbf{Policy Improvement Loop} uses a rubric judge to diagnose recurring failures, construct an adaptive curriculum, and optimize the policy. When progress plateaus and verification becomes a bottleneck, the \textbf{Judge Improvement Loop} selectively queries human guidance on informative failure cases and refines the judge through coactive calibration, in which humans and agents resolve disagreements and converge toward the objective rubric of physical reasoning. The revised judge then guides the next stage of data selection and policy optimization. Experiments on driving and robot navigation tasks demonstrate continuous self-improvement in both policy and judge capability across reinforcement and supervised fine-tuning. These results show how scaling verification supports continuous self-improvement as policy failure patterns evolve. 
\end{abstract}

\begin{document}

\maketitle

\begin{figure}[h]
  \centering
  \vspace{-0.65cm}
  \includegraphics[width=0.98\linewidth]{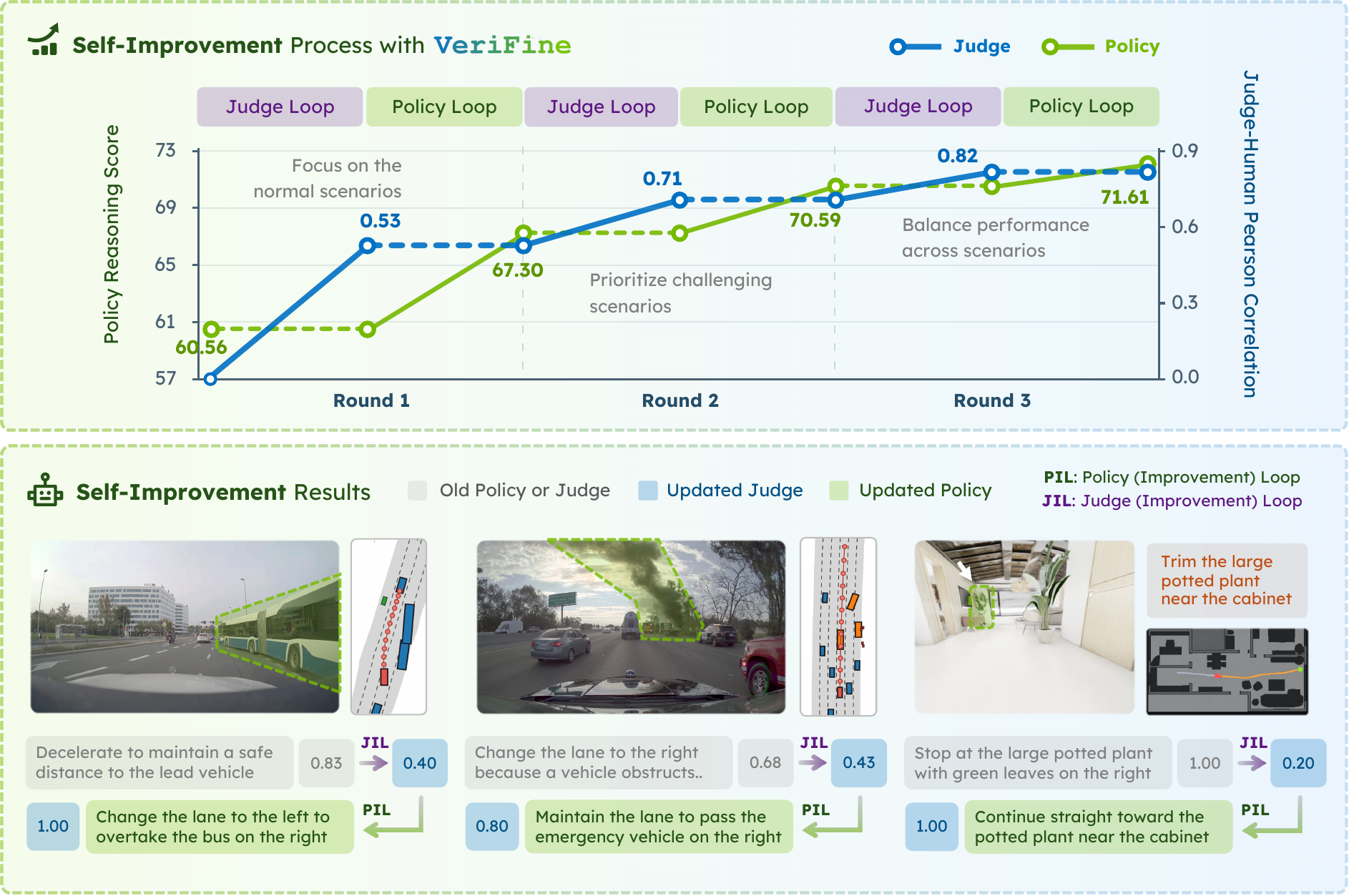}
  \vspace{-0.1cm}
  \captionsetup{font=footnotesize}
  \caption{\VeriFine{} demonstrates continuous self-improvement with the co-evolution of the policy, training curriculum, and judge in an agentic harness framework for embodied reasoning, including autonomous driving and robot navigation tasks. \href{https://veri-fine.github.io/}{\textbf{Project Website}}}
  \label{fig:verifine teaser}
  \vspace{-0.05cm}
\end{figure}

\abscontent

\section{Introduction}
Verification is central to self-improvement: it supplies feedback for policy optimization and determines which training examples are worth learning from \citep{guo2025deepseek,kumar2026llm}. However, policy improvement also changes the judge demands, and the rapid progress of foundation models has increasingly outpaced the ability to evaluate them reliably \citep{gu2026survey,li2025generation}. Most existing approaches rely on a static verification judge, whose capability can quickly become a bottleneck as the policy improves \citep{wang2026co}. Once the policy approaches or exceeds the judge's effective verification boundary, the resulting feedback can become increasingly inaccurate, leading to reward hacking, distribution shift, and even performance degradation \citep{khalaf2026inference}. Meanwhile, static training data also become progressively less informative \citep{koh2026adastar} for further improvement with verification: as previous scenarios are solved, the training distribution becomes dominated by well-solved samples, leaving limited learning signal. This raises a central question for verification: \textit{\uline{How can verification be scaled so that the judge, training curriculum, and policy co-evolve to continuously push the performance frontier?}}

Recent advances in agentic systems and autoresearch \citep{anthropic2025claudecode,openai2025codex} offer a natural framework to realize this co-evolution task by coordinating specialized skills and analyzing failures during iterative optimization. In this work, we study scaling verification within an agentic framework for embodied reasoning, using driving and robot navigation as representative tasks. Unlike conventional language tasks, embodied agents require reasoning in the physical world and handle spatial grounding, temporal understanding, causal reasoning, and safety-aware decision-making \citep{zhang2025safeauto}, making reliable verification challenging. 

Prior work on verification for embodied reasoning focuses on action quality or task completion \citep{sun2026drivejudge, chen2026drivereward}, while evaluating free-form embodied reasoning remains underexplored. Existing reasoning evaluators often rely on constrained question-answering formats and costly human-annotated references \citep{sima2024drivelm}; even LingoJudge \citep{marcu2024lingoqa} only compares against human reasoning without using visual information. Such reference-based evaluation is difficult to scale and sensitive to annotation noise. In contrast, a reference-free judge can assess reasoning directly from physical context, enabling scalable verification and adaptive data selection that filters well-solved scenarios and prioritizes emerging policy failures.

To this end, we study scaling verification as expanding the ability to recognize emerging policy errors and convert them into reusable supervision, and introduce \VeriFine, an agent harness framework for continuous self-improvement through scaling verification. As illustrated in \Cref{fig:verifine overview}, its central mechanism is a pair of coupled improvement loops that connect policy optimization to the evolution of its supervision. The \textbf{Policy Improvement Loop} uses the reference-free judge to identify policy failures and select high-value training samples, tracking the policy's evolving weaknesses. When self-improvement plateaus and the judge approaches its verification boundary, we think little human guidance is necessary, and the \textbf{Judge Improvement Loop} is activated, which selectively queries human guidance on informative failure cases to refine the judge. Rather than treating humans as infallible oracles, we formulate this process as coactive calibration, where humans and agents iteratively resolve ambiguities and converge toward the objective rubric of physical reasoning.

As shown in \Cref{fig:verifine teaser}, we evaluate \VeriFine{} on driving and robot navigation reasoning through reinforcement and supervised fine-tuning (RFT, SFT), respectively. Improvements in both policy and judge across these settings support the role of scaling verification in continuous self-improvement across various tasks or optimization methods. Our main contributions are summarized as follows:

\begin{enumerate}
    \item We introduce \VeriFine{}, an agent-harness framework for continuous self-improvement through scaling verification with the co-evolution of policy, training curriculum, and judge.
    \item We develop a Policy Improvement Loop that leverages judge-identified failure patterns to select high-value data and autonomously optimize the policy without human intervention.
    \item We present a Judge Improvement Loop, which selectively queries human guidance on informative failure cases and performs coactive calibration, enabling the judge to track the evolving policy frontier.
    \item We demonstrate the self-improvement with \VeriFine{} in both policy reasoning and judge capability across driving and robot navigation tasks, establishing the applicability of the framework to reinforcement and supervised fine-tuning.
\end{enumerate}

\section{Related Work}

\subsection{Scaling Verification.} 
Verification assesses the quality or correctness of candidate outputs and provides feedback for selection, monitoring, and policy optimization \citep{kwok2026llm}. At inference time, verification can directly enable test-time scaling with a verifier by allocating additional compute to candidate generation, search, and ranking \citep{snell2025scaling, zhang2025survey}, while the fine-grained critics with the verifier can diagnose errors and guide iterative revision or recovery \citep{gou2024critic,xie2025teaching,wan2026inference}. During training, verification within judges can provide objectives for policy optimization \citep{yuan2024self,liu2025rrm,11092605}, while also filtering noisy samples and prioritizing high-value data to concentrate learning on unresolved failures \citep{zang2025internlmreward,cui2025crpo}. Existing work has begun to construct judge benchmarks \citep{tan2025judgebench,zhou2025evaluating} and explore judge improvement \citep{saha2025learning,whitehouse2026j1}. However, existing work focuses on the improvement of the individual roles of verification \citep{singh2026shelf}, but does not scale verification to close feedback loops among adaptive data selection, policy optimization, judge evolution, and selective human calibration, thereby continuously pushing the performance frontier.

\vspace{-0.1cm}
\subsection{Self-improvement and Autoresearch.} 
Agent-driven self-improvement through hypothesis formation, experimentation, and iterative refinement has emerged as a prominent paradigm, particularly in goal-oriented autoresearch \citep{schmidgall2025agent,chen2026avo}. As self-improvement continuously strengthens the policy model, its verifier must adapt to the resulting capability and distribution
shifts \citep{ding2026evorubrics,xu2026ask}. Recent work has begun to consider the policy and judge improvement together \citep{zhou2026steporlm,zha2026rl} or dynamically adapt the judge to policy behavior \citep{wang2026co,xu2026alternating}. Rubric-based judges have also gained increasing attention because structured criteria provide fine-grained and more accurate judgment beyond a single final score \citep{gunjal2026rubrics,li2026rubricem,liu2026openrubrics}. Nevertheless, autonomous self-improvement eventually encounters bottlenecks when the verifier approaches its effective limit or self-generated feedback becomes unreliable \citep{huang2024large}. At this stage, a small amount of selectively acquired human guidance can become necessary to recalibrate the judge \citep{dwaracherla2024efficient,wang2026reward}. We connect this adaptation to curriculum construction: emerging policy failures guide selective human--agent calibration, and the resulting judge refinements shape subsequent data selection and policy optimization.

\vspace{-0.1cm}
\subsection{Embodied Reasoning.} 
Embodied reasoning requires spatial grounding, temporal understanding, causal reasoning, and safety-aware decision-making \citep{xiong2026phycritic,zhang2025rewind,liu2025robopilot,lee2026roboreward}, making reliable verification challenging. Recent work on driving reasoning has introduced structured reasoning datasets \citep{wang2025alpamayo,huang2026nureasoning} and demonstrated the benefits of reasoning for scene understanding \citep{nie2024reason2drive,coscoy2026mdrive}, generalization \citep{peng2026counterfactual,gerstenecker2026fail2drive}, and planning \citep{zhou2026autovla,zhou2026spanvla}. However, existing evaluators largely use constrained question-answering formats with costly human references \citep{marcu2024lingoqa,wang2024omnidrive}, while driving-specific judges mainly assess trajectory quality \citep{sun2026drivejudge}. In robotics, recent judge models focus on evaluating task progress or completion from video \citep{ma2025vision,tan2025robo,huang2026rynnvalue,liang2026robometer,luo2026mvr} and action-instruction alignment \citep{kwok2026scaling} rather than on high-level free-form reasoning itself. Direct video-grounded verification of free-form embodied reasoning without costly human-annotation references remains underexplored.

\begin{figure}[t]
  \centering
  \includegraphics[width=0.99\linewidth]{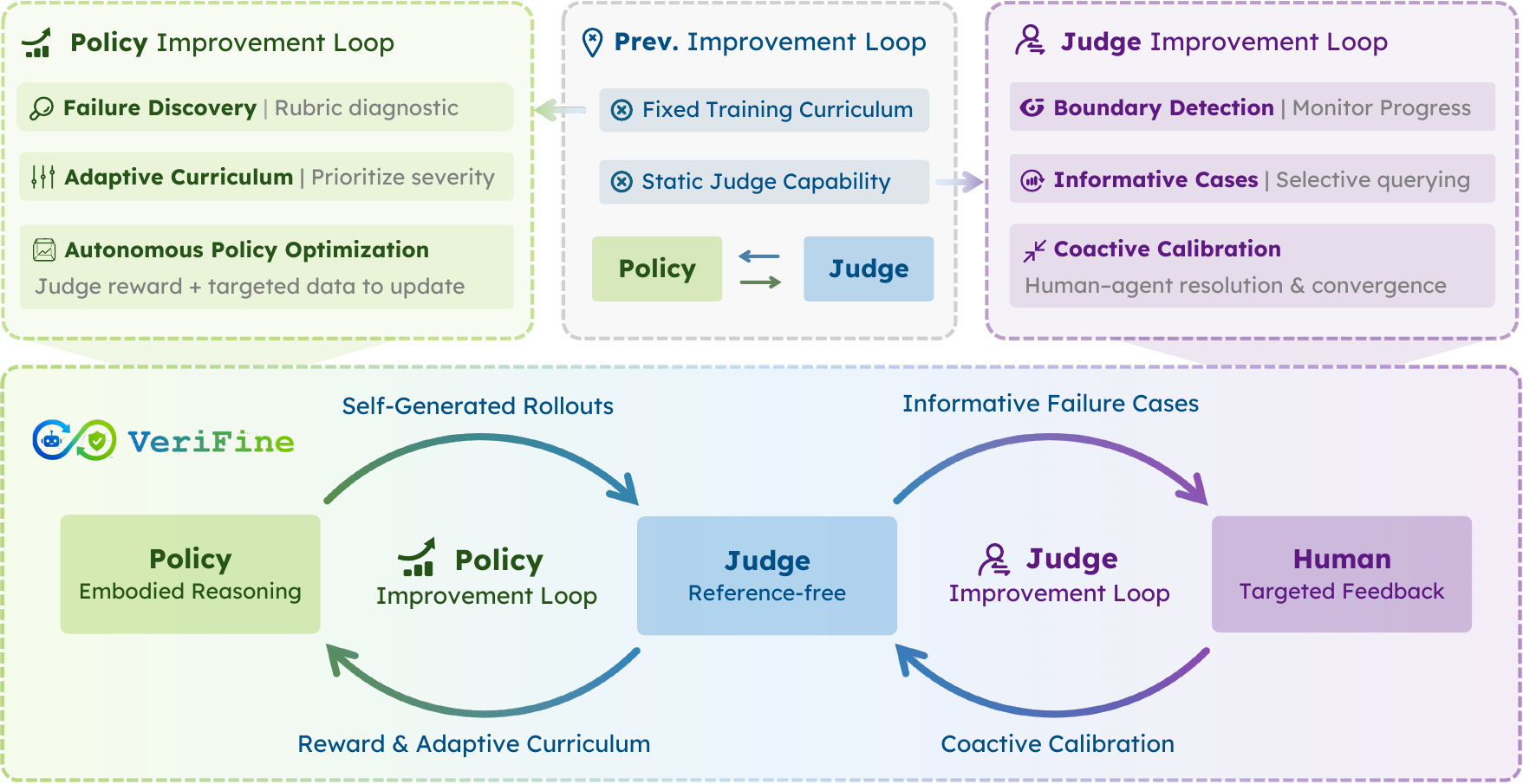}
  \caption{\VeriFine{} is an agent harness framework for scaling verification. The \textcolor[HTML]{275317}{\textbf{Policy Improvement Loop}} uses a reference-free judge to diagnose recurring failures, adapt the training curriculum, and optimize the policy. When verification limits further progress, the \textcolor[HTML]{7030A0}{\textbf{Judge Improvement Loop}} selectively acquires human guidance and refines the judge through coactive calibration, resolving disagreements in rubric-based judgments. The updated judge guides subsequent data selection and policy optimization, allowing verification capability to evolve with the policy's failure patterns.}
  \label{fig:verifine overview}
  \vspace{-0.15cm}
\end{figure}

\section{VeriFine}
\label{sec:method}
We formulate scaling verification as an iterative co-evolution process among the policy, training curriculum, and reference-free judge, as illustrated in \Cref{fig:verifine overview}.

\subsection{Problem Formulation and Framework Overview}
\label{sec:formulation}
Let $x_i$ denote the observable physical context of sample $i$, including visual observations, ego states, and task instructions. At iteration $t$, a policy $\pi_{\theta_t}$, parameterized by $\theta_t$, generates a task output $\hat{y}_i\sim\pi_{\theta_t}(\cdot\mid x_i)$, which contains both free-form reasoning and task-specific decisions. In our instantiation, $\pi_{\theta_t}$ is a vision-language-action (VLA) model with $\hat{y}_i=(\hat{r}_i,\hat{\mathbf{a}}_i)$, where $\hat{r}_i$ is the generated reasoning and $\hat{\mathbf{a}}_i$ is the predicted action sequence. A judge $J_{\phi_t}$, parameterized by $\phi_t$, evaluates the policy output against the physical context $x_i$. Starting from an initial policy $\pi_{\theta_0}$, judge $J_{\phi_0}$, and large-scale candidate data pool $\mathcal{U}$, in the Policy Improvement Loop, the agent harness coordinates failure analysis, data selection, training, and evaluation through specialized tools. The training curriculum $\mathcal{D}_t$ is constructed and the next policy is updated as:
\begin{equation}
    \mathcal{D}_t =
    \operatorname{CurriculumSelect}
    (\mathcal{U},\pi_{\theta_t},J_{\phi_t}),
    \qquad
    \theta_{t+1}
    =
    \operatorname{PolicyUpdate}
    (\theta_t,\mathcal{D}_t,J_{\phi_t}).
    \label{eq:policy_loop}
\end{equation}
When verification becomes the primary bottleneck, the Judge Improvement Loop selectively acquires human guidance and coactively calibrates to refine the judge:
\begin{equation}
    \mathcal{H}_t =
    \operatorname{HumanQuery}
    (\mathcal{U},\pi_{\theta_{t+1}},J_{\phi_t}),
    \qquad
    \phi_{t+1} =
    \operatorname{JudgeUpdate}
    (\phi_t,\mathcal{H}_t).
    \label{eq:judge_loop}
\end{equation}

\VeriFine{} maintains supervision state across rounds, including the revised judge rubric and curriculum construction. Each policy's observed failures inform subsequent updates to this state. In our controlled experiments, we train each new policy from the same base initialization, allowing us to compare successive supervision states under a fixed per-policy training budget. System-level improvement therefore accumulates through the verification and curriculum process across rounds.

\subsection{Reference-Free Rubric Judge as the Verification Substrate}
\label{sec:reference_free_judge}

The judge serves as the verification substrate and a shared interface to connect policy optimization, curriculum construction, and human guidance. Unlike a reference-based judge $J_{\phi}(\hat{y}_i,y_i^{\mathrm{ref}})$, our judge evaluates the outputs directly from the physical context without human-annotated reasoning:
\vspace{0.01cm}
\begin{equation}
J_{\phi_t}(x_i,\hat{y}_i)
=
(s_i,\mathbf{z}_i,e_i),
\vspace{0.1cm}
\end{equation}
where $s_i$ is the overall quality score for the sample $i$, $\mathbf{z}_i$ are the structured diagnostic scores, and $e_i$ is an optional evaluation explanation. We name the judge \textbf{VeriFine-Judge} in this paper.

To reliably evaluate free-form reasoning, we instantiate the judge with a frontier vision-language model (VLM) and an explicit rubric that produces structured diagnostic scores $\mathbf{z}_i$ rather than a single opaque score. However, frontier VLMs are too costly and slow during large-scale policy optimization such as in reinforcement fine-tuning. We therefore adopt a teacher--student design: a frontier VLM serves as the teacher, whose rubric is coactively calibrated with humans as described in \Cref{sec:coactive_calibration}, and its evaluation capability is then distilled into a compact VLM student to balance verification quality with efficiency. More implementation details are provided in \Cref{app:judge}.

The rubric organizes embodied reasoning into two groups of criteria: \emph{Action} and \emph{Component}. Taking driving as an example, Action evaluates instruction consistency and safety, while Component evaluates visual grounding, coverage of behavior-relevant scene elements, and their causal relationship to the proposed action. Task-specific dimensions and score aggregation are detailed in \Cref{app:judge}.

\subsection{Policy Improvement Loop}
\label{sec:policy_loop}
This loop converts judge-identified failure patterns into training decisions through multiple rounds of autonomous data selection, policy optimization, and evaluation.

\subsubsection{Failure Discovery and Adaptive Curriculum}
\label{sec:failure_discovery} 
At iteration $t$, the current policy generates outputs over an anchor policy evaluation subset $\mathcal{P}$. This subset is quality-checked and selected by human experts to ensure data reliability and broad coverage of diverse scenario types, including both reasoning-dense and reasoning-light cases. Further details on $\mathcal{P}$ are provided in \Cref{app:policy eval set}. The judge then evaluates the policy outputs and produces structured evaluation results:
\vspace{0.01cm}
\begin{equation}
\mathcal{E}_t
=
\left\{
(x_i,\hat{y}_i,s_i,\mathbf{z}_i)
\right\}_{i\in\mathcal{P}}.
\vspace{0.1cm}
\end{equation}
The \VeriFine{} agent aggregates these evaluations across scenarios and rubric dimensions to identify recurring failure patterns, distinguishing systematic errors from isolated failures. 

These patterns are translated into executable selection criteria over $\mathcal{U}$. Selection serves two distinct purposes: prioritizing contexts in which the policy needs improvement and filtering unreliable training targets. A scenario that exposes a policy weakness can be valuable for learning, while an incorrect response to that scenario is unsuitable as a supervised demonstration. The curriculum is revisited as policy weaknesses and judge change, allowing a revised judge to recover useful supervision or exclude examples that were previously misjudged.

\subsubsection{Judge-Guided Policy Optimization}

The selected curriculum $\mathcal{D}_t$ supports either reinforcement or supervised fine-tuning. For the driving reasoning task, we instantiate the update with reinforcement fine-tuning using GRPO, and the overall judge score supplies the reward to be maximized:
\begin{equation}
\mathcal{R}_{\text{RFT}}^{\text{Driving}}(\theta)=
\mathbb{E}_{
x_i\sim\mathcal{D}_t,\,
\hat{y}_i\sim\pi_{\theta}(\cdot\mid x_i)
}
\left[
s_{\phi_t}(x_i,\hat{y}_i)
-
\beta D_{\mathrm{KL}}
\left(
\pi_{\theta}(\cdot\mid x_i)
\,\Vert\,
\pi_{\theta_{\mathrm{ref}}}(\cdot\mid x_i)
\right)
\right],
\end{equation}
where $\pi_{\mathrm{ref}}$ is the reference policy, and the KL-divergence term regularizes the updated policy against excessive deviation from the current policy. For navigation, we instantiate the update with supervised fine-tuning, and the judge selects high-value and high-quality demonstrations to minimize:
\begin{equation}
\mathcal{L}_{\text{SFT}}^{\text{RobNav}}(\theta)
=-\mathbb{E}_{(x_i,y_i^\star)\sim\mathcal{D}_t}
\left[
\log\pi_{\theta}(y_i^\star\mid x_i)
\right],
\end{equation}
where $y_i^\star$ denotes the target data from the selected curriculum $\mathcal{D}_t$. Here, judge scores determine training-set membership rather than directly entering the optimization objective. Then, candidate policies are assessed under a consistent monitoring protocol. The resulting failure analysis guides subsequent curriculum updates and identifies potential limitations of the current judge.

\subsection{Judge Improvement Loop}
This loop converts informative policy failures into reusable improvements in verification, enabling further learning when the current judge becomes a bottleneck.

\subsubsection{Boundary Detection and Human Query Selection}
We monitor policy progress on the anchor policy evaluation set $\mathcal{P}$. Thus, rather than using the dynamically evolving reference-free judge model as the evaluator, we develop an anchor reference-based judge model, VeriFine-Judge-RB, based on high-quality human annotations. Let $P_t$ denote the aggregate performance of $\pi_{\theta_t}$ on $\mathcal{P}$, $\Delta P_t=P_t-P_{t-1}$, and $\Delta R_t^{\mathrm{judge}}$ the corresponding performance change. We measure performance stagnation and judge--performance discrepancy as
\begin{equation}
    \overline{\Delta P}_t
    =
    \frac{1}{K}
    \sum_{k=0}^{K-1}
    \Delta P_{t-k},
    \quad
    G_t
    =
    \left|
        \Delta R_t^{\mathrm{judge}}
        -
        \Delta P_t
    \right|,
    \label{eq:verification_progress}
\end{equation}
where ${K}$ denotes the number of most recent policy changes considered, a small $\overline{\Delta P}_t$ indicates that improvement has plateaued, while a large $G_t$ indicates that increasing judge reward no longer corresponds to improved performance on $\mathcal{P}$. The trigger rule, thresholds, and check frequency are specified in \Cref{app:loop_protocol}.

Once activated, the agent constructs the current judge evaluation set $\mathcal{H}_t$ containing cases with uncertain, inconsistent, and high-impact evaluations. Importantly, $\mathcal{H}_t$ is constructed from outputs rolled out by the newly updated policy $\pi_{\theta_{t+1}}$. It can track the policy's current failure pattern. Moreover, judge refinement deliberately retains both positive and negative outputs. The low-scoring cases are important for defining the judge's decision boundary and calibrating its responses to policy errors.

\subsubsection{Coactive Calibration}
\label{sec:coactive_calibration}

The Judge Improvement Loop receives the current failure-pattern analysis and the selected evaluation set $\mathcal{H}_t$. The agent uses them to update a cumulative judge evaluation set:
\begin{equation}
    \mathcal{J}_t
    =
    \mathcal{J}_{t-1}
    \cup
    \mathcal{H}_t,
    \qquad
    \mathcal{J}_{-1}
    =
    \varnothing.
    \label{eq:judge_evaluation_set}
\end{equation}
Its cumulative construction preserves cases from previous iterations, preventing new refinements from degrading previously acquired verification capabilities. During initial judge construction, the same procedure is based on the base policy $\pi_{\theta_0}$. Further details are provided in \Cref{app:judge eval set}.

Human experts first provide a coarse calibration of the judge scores on $\mathcal{J}_t$ by inspecting the physical context and candidate outputs. Then, the agent launches an autonomous process that iteratively proposes, evaluates, and selects revisions to the rubric. When alignment with the current human calibration plateaus, the agent presents the disputed cases together with its structured diagnostic scores and evaluation explanations. Human experts may confirm the evaluation, correct individual rubric dimensions, clarify underspecified criteria, or even revise their previous judge scores annotation in evaluation set. Human judgments are not treated as infallible ground truth. Instead, the agent and human experts repeatedly expose disagreements and refine their interpretations until no further alignment gain can be achieved, converging toward the shared latent rubric of embodied reasoning.

\section{Experiments}

We evaluate \VeriFine{} through four questions:

\textbf{Q1}: \textit{Can the policy and judge improve together as policy failure patterns evolve?}

\vspace{-0.12cm}
\textbf{Q2}: \textit{How does each improvement loop contribute to the resulting policy and judge?}

\vspace{-0.12cm}
\textbf{Q3}: \textit{Does the evolved judge generalize and provide more effective verification?}

\vspace{-0.12cm}
{\hangindent=1.9em
\hangafter=1
\noindent
\textbf{Q4}: \textit{Does the framework support self-improvement across different embodied reasoning tasks and optimization methods?}
\par}

\begin{table*}[t]
\centering

\caption{
\textbf{Comparison of Different Judges in the Policy Improvement Loop} with a common initialization and matched per-policy training budgets.The curriculum and reward judges refer to the judges for data selection and policy optimization, respectively; reference-free variants use the same for both.
}

\label{tab:policy_improvement}

\fontsize{8pt}{9pt}\selectfont
\setlength{\tabcolsep}{14pt}
\renewcommand{\arraystretch}{1.4}

\begin{tabular}{l l | c c | c c}
\toprule
\makecell[l]{Curriculum\\Judge}
&
\makecell[l]{Reward\\Judge}
&
\makecell{Reasoning Score\\(RB) $\uparrow$}
&
\makecell{Reasoning Score\\(RF) $\uparrow$}
&
\makecell{minADE$_6$\\(m) $\downarrow$}
&
\makecell{ADE\\(m) $\downarrow$}
\\
\midrule

\rowcolor[HTML]{E5DDEC}
\multicolumn{6}{l}{\textit{Base Policy Model}} \\
None
& None
& 60.56
& 68.05
& 1.049
& 2.139
\\

\midrule
\rowcolor[HTML]{E8EDF3}
\multicolumn{6}{l}{\textit{Reference-Based Judges}} \\
Random
& LingoJudge
& 63.45
& 69.82
& 1.064
& 2.133
\\

LingoJudge
& LingoJudge
& 65.75
& 72.60
& 1.172
& 2.242
\\

Random
& VeriFine-Judge-RB
& 64.03
& 69.95
& 1.080
& 2.166
\\

VeriFine-Judge-RB
& VeriFine-Judge-RB
& 67.70
& 75.20
& 1.164
& 2.235
\\

\midrule
\rowcolor[HTML]{EDF4DB}
\multicolumn{6}{l}{\textit{Reference-Free Judges}} \\

\multicolumn{2}{l|}{
  VeriFine-Judge-\textsc{R1} (\textsc{First Round})
}
& 67.30
& 77.51
& 1.106
& 2.177
\\

\multicolumn{2}{l|}{
  VeriFine-Judge-\textsc{R2} (\textsc{Second Round})
}
& 70.59
& 79.61
& 1.078
& 2.139
\\

\multicolumn{2}{l|}{
  VeriFine-Judge-\textsc{R3} (\textsc{Third Round})
}
& \textbf{71.61}%
  \makebox[0pt][l]{%
    \hspace{0.35em}%
    {\scriptsize\textcolor{nvgreen}{+18.2\%}}%
  }
& \textbf{83.13}%
  \makebox[0pt][l]{%
    \hspace{0.35em}%
    {\scriptsize\textcolor{nvgreen}{+22.2\%}}%
  }
& \textbf{1.029}
& \textbf{2.117}
\\

\bottomrule
\end{tabular}
\vspace{-0.18cm}
\end{table*}

\subsection{Main Experimental Setup}
We conduct three self-improvement rounds on driving reasoning. Unless otherwise specified, the following settings refer to this task; detailed navigation settings are provided in the appendix.

\vspace{0.08cm}
\textbf{Driving Reasoning Data.}
Our main experiments use a large-scale internal driving-reasoning dataset covering 2 million clips of human driving across 25 countries, with synchronized multi-view videos, trajectories, and free-form reasoning. We construct a fixed anchor policy evaluation set $\mathcal{P}$ of 2,134 samples and an independent policy test set $\mathcal{P}^{\mathrm{test}}$ of 2,772 samples. Those anchor sets are manually quality-checked to ensure reliable annotations and broad coverage of scenario types. Judge calibration uses incrementally selected nominal and challenging scenarios. Across the first two judge-improvement rounds, the selected judge evaluation subsets $\mathcal{H}_1$ and $\mathcal{H}_2$ are constructed, containing 433 normal and 179 challenging driving scenarios. The corresponding judge test sets are $\mathcal{H}_1^\mathrm{test}$ and $\mathcal{H}_2^\mathrm{test}$ with 514 and 218 scenarios. The final judge test set $\mathcal{J}^\mathrm{test}$ is their union. Test labels are frozen and excluded from rubric refinement. Details are provided in \Cref{app:eval details}.

\vspace{0.08cm}
\textbf{Policy and Judge Models.}
\label{sec:driving reasoning setup}
We initialize the driving policy of the Alpamayo 1.5 8B \citep{nvidia2026alpamayo1.5} model. Policy optimization is performed using GRPO \citep{shao2024deepseekmath} with the overall reference-free judge score as the reward. The judge follows the teacher--student design introduced in \Cref{sec:reference_free_judge}. We use Claude Opus 5 \citep{anthropic2026claudeopus5} as the base model for the rubric teacher and distill its evaluation capability into a compact model after coactive calibration. The student model is built on a Qwen3-VL 2B backbone \citep{bai2025qwen3vl}, and we pretrain the backbone to have a better vision encoder on our internal driving dataset with the VLA planning task. More details of the policy and judge model are provided in \Cref{app:judge}.

\vspace{0.08cm}
\textbf{Baselines.}
We compare the base policy with RFT using random or judge-guided data selection with 2,700 optimization steps on 25K examples selected from a shared 400K pool. LingoJudge \citep{marcu2024lingoqa} and our VeriFine-Judge-RB serve as alternative reference-based judges. We also report \VeriFine{} after each self-improvement round (R1--R3). Each round denotes a fixed-judge policy improvement stage, including curriculum updates until the prescribed stopping criterion is met. Each round for driving and robot navigation reports stage-end policies, each trained from the same base checkpoint with matched training-set sizes and optimization budgets, and more details are in \ref{app:implementation}.

\begin{figure}[t]
  \centering
  \includegraphics[width=0.99\linewidth]{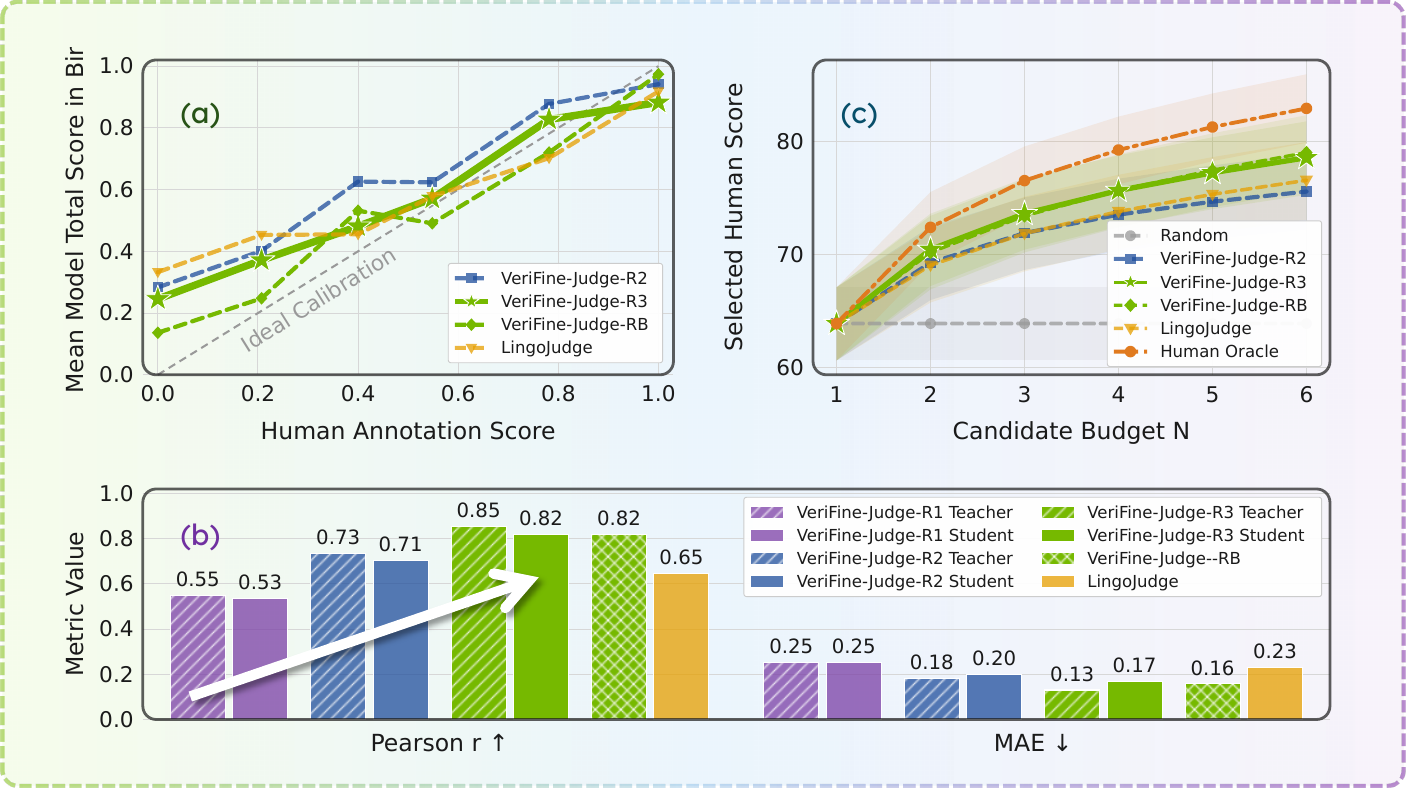}
  \caption{\textbf{Judge Performance Comparison.} \textcolor[HTML]{275317}{\textbf{(a) Calibration curves}} showing the mean judge score within each human-score bin; the gray diagonal denotes ideal calibration. \textcolor[HTML]{7030A0}{\textbf{(b) Judge--human alignment}} measured by Pearson correlation and MAE for teacher and student judges across improvement rounds and reference-based baselines. \textcolor[HTML]{084F6A}{\textbf{(c) Test-time scaling}} is evaluated by the human score of the selected candidate as the candidate budget increases. Shaded regions denote 95\% bootstrap CI.}
  \label{fig:judge results}
\end{figure}

\vspace{0.08cm}
\textbf{Evaluation Metrics.}
All policies are evaluated on $\mathcal{P}^{\mathrm{test}}$ using the fixed reference-based evaluator, VeriFine-Judge-RB. We additionally report scores from the same final reference-free judge for all policies. Policy metrics include reasoning scores on a 0--100 scale, ADE, and $\mathrm{minADE}_6$. Judge quality is measured by Pearson correlation $r$ and MAE against human scores, with MAE normalized to a 0--1 scale.  More details are in \Cref{app:exp}. Besides, on the judge test set, VeriFine-Judge-RB achieves $r$ 0.82, compared with 0.65 for the SOTA method LingoJudge, as shown in \Cref{fig:judge results}. This provides empirical support for the fixed evaluator used to compare policies across rounds.

\subsection{Self-Improvement Results with Scaling Verification}

\vspace{0.25cm}
\textbf{Judge--Policy Co-Evolution (Q1).} 
\Cref{fig:verifine teaser} summarizes the coupled improvement process in driving reasoning task, whose focus progresses from nominal driving scenarios to challenging failures and then to balanced refinement. Across three rounds, the policy reasoning score increases from 60.56 to 71.61 under the fixed reference-based evaluator, an 18.2\% relative improvement over the base policy. Meanwhile, student judge--human correlation reaches 0.82 on the final judge test set, indicating strong alignment after coactive calibration. These results reveal a complementary cycle: policy improvement exposes new verification limitations, while judge refinement restores reliable supervision for further policy improvement.

\textbf{Policy Improvement Loop (Q2).}
\Cref{tab:policy_improvement} examines the use of judges for curriculum construction and policy optimization. With VeriFine-Judge-RB held fixed as the reward, judge-guided selection improves the reasoning score from 64.03 to 67.70 over random selection, demonstrating the value of directing training toward selected examples. The first reference-free round reaches 67.30, comparable to the strongest reference-based configuration (67.70) without any costly annotation reliance during inference. Under the same policy-training budget, subsequent rounds reach 70.59 and 71.61, with R3 exceeding the strongest reference-based baseline by 5.8\%. The final policy also achieves the lowest ADE, even without any action reward in training. With each round's reward judge held fixed, judge-guided selection improves RB performance over random selection by 4.5--14.6\% (Appendix~\Cref{tab:abl curriculum}). These results demonstrate a stronger policy yield from the evolved curriculum and judge in the self-improvement loop. 

Using the final VeriFine-Judge as a fixed reference-free evaluator yields a similar improvement trend across rounds, with a relative gain of 22.2\% compared with 18.2\% under a reference-based evaluator, supporting its effectiveness as an evaluation metric. The embodied decisions admit multiple plausible futures; the reference-free judge can credit grounded, safe alternatives that differ from the recorded behavior, yielding a more permissive assessment without requiring reference agreement.

\textbf{Judge Improvement Loop (Q2).}
\Cref{fig:judge results} (a,b) shows progressively stronger judge--human alignment: teacher correlation increases from 0.55 to 0.85, while MAE decreases from 0.25 to 0.13. The final student achieves $r=0.82$ and MAE $=0.17$, comparable to the reference-based judge VeriFine-Judge-RB (0.82/0.16) and better than LingoJudge (0.65/0.23) without requiring reference reasoning. Moreover, with the Opus~5 backbone held fixed, coactive calibration improves correlation from 0.67 to 0.85 (Appendix \Cref{fig:teacher model comparison}), demonstrating gains from refinement beyond backbone capability. Student architecture ablations and latency performance are reported in Appendix \Cref{tab:student_judge_variants}.

The failure-specific results reveal how the loop expands the verification boundary. As policy failure analysis shifts attention toward challenging scenarios, the loop identifies errors in the current judge's assessments and targets them through coactive calibration. Our R1 student achieves $r=0.72$ on the first-round test subset but only $r=0.07$ on the later challenging subset; refinement raises correlation on the latter to $r=0.74$ in R2 (Appendix~\Cref{fig:judge results of subsets}). R3 then targets remaining errors across the accumulated scenarios without adding a new test subset. Through this diagnosis--calibration cycle, newly exposed verification errors become reusable judge refinements that inform subsequent curriculum construction and policy optimization.

\subsection{Test-Time Scaling with Evolved Judge (Q3)}
To assess verification beyond the training loop, a separate policy generates six reasoning--action candidates for each of 401 independent scenarios. All judges select from the same candidate pool as the sampling budget increases from one to six. \Cref{fig:judge results} (c) compares the R2 and R3 judges with random selection, LingoJudge, VeriFine-Judge-RB, and a human-oracle upper bound.

At a budget of six candidates, our final judge selects outputs with a human-rated reasoning score of 78.6, outperforming the R2 judge and LingoJudge while performing comparably to VeriFine-Judge-RB. Because candidate generation is held fixed, the gains reflect improved selection rather than a stronger generating policy. These results demonstrate effective generalization of the evolved judge.

\begin{figure}[t]
  \centering
  \includegraphics[width=0.99\linewidth]{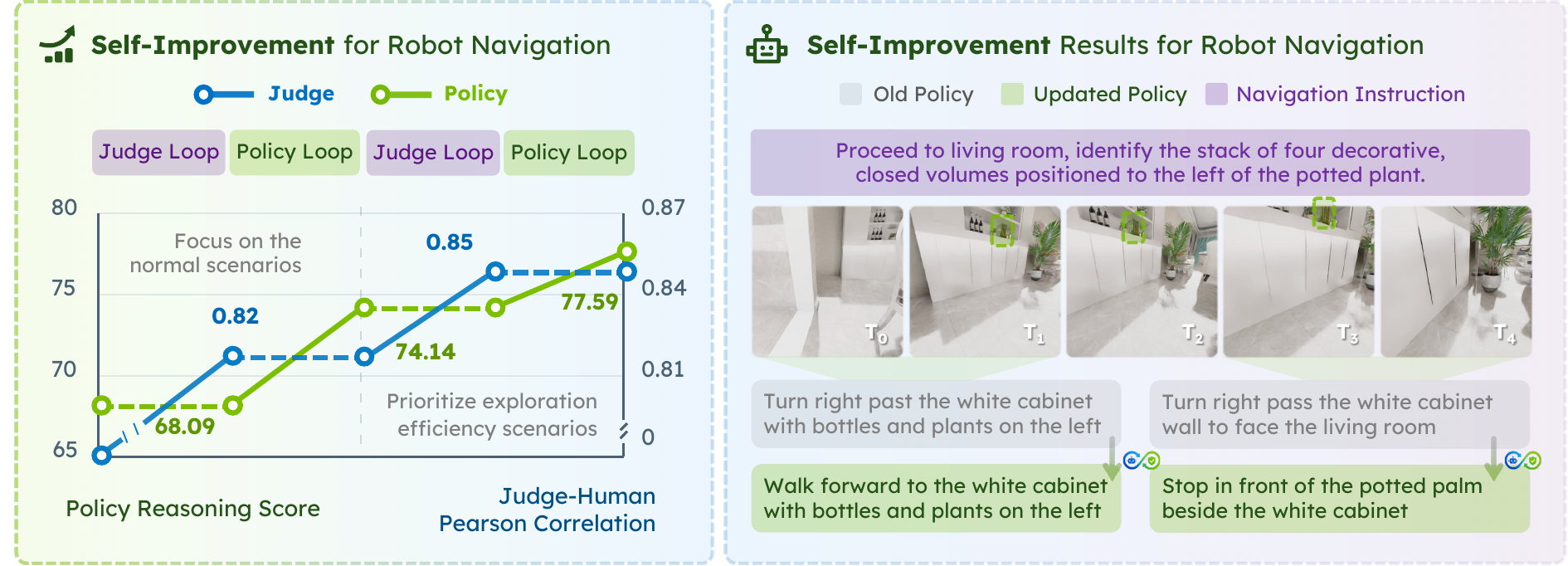}
  \caption{\textbf{Self-improvement on Robot Navigation Task.} The left part shows the quantitative results in the robot navigation task with significant improvement in final policy performance, and the right part shows the qualitative results with a sequential reasoning example: the updated policy with \VeriFine{} corrects the action failures of the old policy.}
  \label{fig:nav results}
\end{figure}

\subsection{Framework Generalization to Robot Navigation}

\vspace{0.25cm}
\textbf{Robot Navigation Task Setup.}
We instantiate \VeriFine{} on robot navigation using data collected in VLNVerse scenarios \citep{lin2025vlnverse} and a Qwen3-VL 2B policy. Unlike driving reasoning tasks, this policy optimization uses supervised fine-tuning: the judge only contributes to the training curriculum rather than supplying an RL reward. We retain the two-loop framework with a navigation-specific rubric. Dataset, training, and evaluation details are provided in the appendix.

\textbf{Cross-Task Self-Improvement (Q4).}
\Cref{fig:nav results} shows improvements in both policy reasoning and judge alignment as the optimization focus shifts from nominal scenarios toward exploration efficiency. The final policy reasoning score improves by 14\% from 68.09 to 77.59, while the final judge--human correlation approaches 0.85. The qualitative example in \Cref{fig:nav results} and \Cref{fig:verifine teaser} illustrates corrected movement, goal understanding, and stopping decisions after our self-improvement framework. These results extend the role of evolving verification beyond reward-based optimization: it can also support self-improvement by refining the supervision selected for SFT. Moreover, the resumed policy gains after judge refinement mirror the driving results, demonstrating that the co-evolution framework transfers across embodied-reasoning tasks.

\section{Conclusions}
We introduced \VeriFine{}, an agent-harness framework for continuous self-improvement through scaling verification with the co-evolution of policy, training curriculum, and judge. Its Policy Improvement Loop adapts the training curriculum to recurring failures and optimizes the policy, while its Judge Improvement Loop detects verification bottlenecks and uses selective human guidance for coactive calibration. A reference-free judge provides the structured feedback connecting these two loops. Experiments on driving and robot navigation demonstrate improvements in both policy and judge across reinforcement and supervised fine-tuning. These results show how scaling verification supports continuous self-improvement as policy failure patterns evolve.

\clearpage
\setcitestyle{numbers}
\bibliographystyle{plainnat}
\bibliography{main}

\clearpage
\appendix

\section[VeriFine Appendix]{%
  \texorpdfstring{%
    \raisebox{-.16\height}{%
      \includegraphics[width=0.06\textwidth]{fig/logo.png}%
    }%
    \hspace{0.2em}%
    \textbf{%
      \protect\gradientRGB{VeriFine}{1,111,214}{121,199,5}%
    }%
    \hspace{0.25em}Appendix%
  }{VeriFine Appendix}%
}

\vspace{0.5cm}
\begingroup
\linespread{1.6}\selectfont

\etocsetnexttocdepth{subsubsection}

\makeatletter

\let\vfOriginalSubsectionLine\l@subsection
\let\vfOriginalSubsubsectionLine\l@subsubsection

\renewcommand{\l@subsection}[2]{%
  \hypersetup{linkcolor=nvgreen}%
  \vfOriginalSubsectionLine{#1}{#2}%
}

\renewcommand{\l@subsubsection}[2]{%
  \hypersetup{linkcolor=black}%
  \vfOriginalSubsubsectionLine{#1}{#2}%
}

\makeatother

\etocsettocstyle{}{%
  \vspace{1em}
}

\localtableofcontents

\endgroup

\subsection{Additional Experiment Results}
\label{app:results}

We provide additional analyses of the two improvement loops. We first examine reward supervision and curriculum construction, followed by teacher-backbone and student-architecture comparisons and an analysis of judge performance across evolving failure distributions.

\subsubsection{Additional Results on Policy Improvement Loop}

\VeriFine{} targets system-level self-improvement through the coordinated evolution of the policy, judge, and curriculum. Each curriculum is constructed from policy failures diagnosed by the current judge, making it an outcome of the feedback process rather than an independently specified component. Cross-round judge--curriculum swaps therefore assess component transfer across stages rather than an alternative execution of the self-improvement loop. We evaluate system-level gains under matched per-policy training budgets and examine component contributions through within-stage comparisons that hold the reward judge fixed while varying data selection.

\vspace{0.15cm}
\textbf{Ablation of Reward Supervision and Curriculum Construction.}
\Cref{tab:policy_improvement} ablates the contributions of reward supervision and curriculum construction.

\textit{(i) Reward supervision.}
Under the fixed RB evaluator, RFT on randomly selected data with LingoJudge and VeriFine-Judge-RB improves reasoning scores by 4.8\% and 5.7\%, respectively, over the base policy, demonstrating useful but limited gains from reward-based optimization alone.

\textit{(ii) Judge-guided curriculum.}
Holding each reward judge fixed, replacing random selection with the corresponding judge-guided curriculum yields further relative improvements of 3.6\% and 5.7\%, demonstrating the benefit of directing training toward informative examples under unchanged reward supervision.

\textit{(iii) Policy--curriculum--judge co-evolution.}
With our reference-free judge, the first round already achieves performance comparable to the strongest reference-based configuration. As new failure patterns emerge, \VeriFine{} refines verification and updates the curriculum, enabling continued policy gains across subsequent rounds. Together, these results support coordinated refinement of verification and curriculum construction under a fixed per-policy training budget.

\textbf{Ablation of Curriculum Construction of VeriFine.}
\Cref{tab:abl curriculum} compares random and judge-guided data selection while holding each round's reward judge fixed, with matched initialization, training-set size, and optimization budget. Judge-guided selection improves RB reasoning scores over random selection by 14.6\%, 8.1\%, and 4.5\% in R1, R2, and R3, respectively. The RF evaluator shows the same ordering. The curriculum advantage persists across judge versions, supporting the use of evolving verification both to assess policy outputs and to identify useful training examples.

\begin{table*}[t]
\centering

\caption{
\textbf{Curriculum Ablation across Judge Improvement Rounds.}
        Within each round, random and judge-guided selection share the same reward judge, base-policy initialization, and policy-training budget. Shaded rows use judge-guided selection. All policies are evaluated using the same fixed RB and final RF evaluators.
}
\label{tab:abl curriculum}

\fontsize{8pt}{9pt}\selectfont
\setlength{\tabcolsep}{16pt}
\renewcommand{\arraystretch}{1.9}

\begin{tabular}{l l | c c | c c}
\toprule
\makecell[l]{Curriculum\\Judge}
&
\makecell[l]{Reward\\Judge}
&
\makecell{Reasoning Score\\(RB) $\uparrow$}
&
\makecell{Reasoning Score\\(RF) $\uparrow$}
&
\makecell{minADE$_6$\\(m) $\downarrow$}
&
\makecell{ADE\\(m) $\downarrow$}
\\
\midrule

Random
& VeriFine-Judge-\textsc{R1}
& 58.71
& 68.90
& 1.349
& 2.594
\\

\rowcolor[HTML]{EDF4DB}
\multicolumn{2}{l|}{
  VeriFine-Judge-\textsc{R1} (\textsc{First Round})
}
& 67.30
& 77.51
& 1.106
& 2.177
\\
\midrule

Random
& VeriFine-Judge-\textsc{R2}
& 65.32
& 76.67
& 1.119
& 2.317
\\

\rowcolor[HTML]{EDF4DB}
\multicolumn{2}{l|}{
  VeriFine-Judge-\textsc{R2} (\textsc{Second Round})
}
& 70.59
& 79.61
& 1.078
& 2.139
\\
\midrule

Random
& VeriFine-Judge-\textsc{R3}
& 68.53
& 81.59
& 1.053
& 2.222
\\

\rowcolor[HTML]{EDF4DB}
\multicolumn{2}{l|}{
  VeriFine-Judge-\textsc{R3} (\textsc{Third Round})
}
& {71.61}
& {83.13}
& {1.029}
& {2.117}
\\

\bottomrule
\end{tabular}
\end{table*}

\subsubsection{Additional Results on Judge Improvement Loop}
\label{app:judge results}

\begin{figure}[t]
  \centering
   \includegraphics[width=\linewidth]{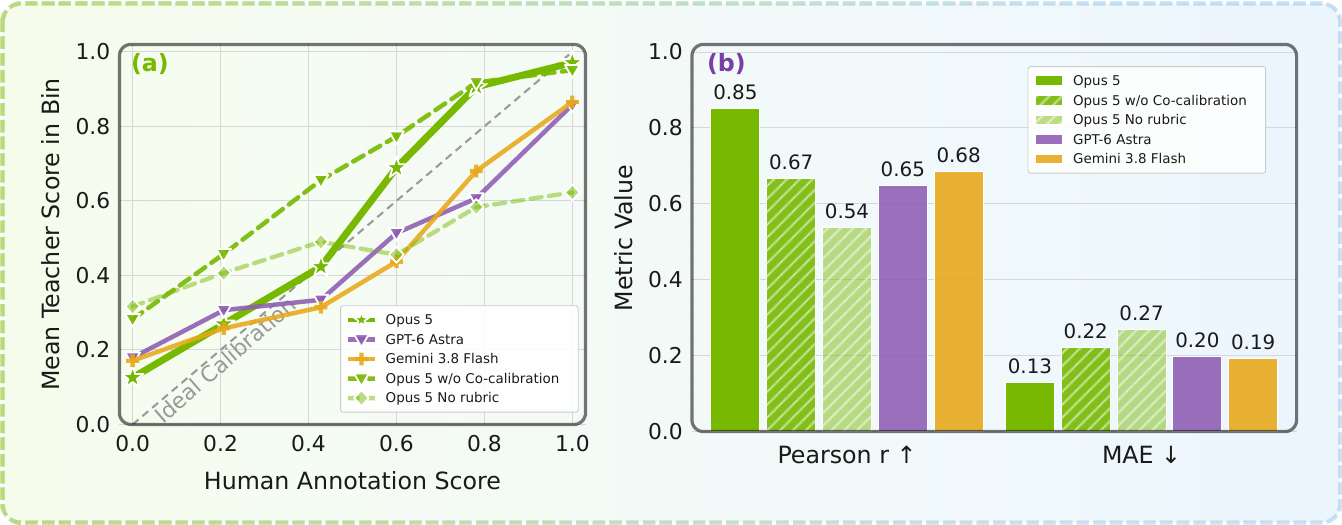}
  \caption{\textbf{Comparison of Backbone Variants of the Teacher Judge Model}. All the rubric variants, except the one without coactive calibration, use the same rubric, but with different backbone models. No rubric variant uses a straightforward instruction without detailed rubrics for judgment. \textcolor[HTML]{275317}{\textbf{(a) Calibration curves}} showing the mean judge score within each human-score bin; the gray diagonal denotes ideal calibration. \textcolor[HTML]{7030A0}{\textbf{(b) Judge--human alignment}} measured by Pearson correlation and MAE for backbone variants.}
  \label{fig:teacher model comparison}
\end{figure}

\vspace{0.25cm}
\textbf{Teacher Judge Model Backbone.}
\Cref{fig:teacher model comparison} compares teacher-judge backbones under a controlled setting in which all rubric-based variants use the same final rubric. Across backbones, rubric-based judging consistently outperforms judging without an explicit rubric, demonstrating the importance of structured evaluation criteria. Coactive calibration further substantially improves the Opus~5 judge \citep{anthropic2026claudeopus5} over its pre-calibration counterpart, showing that the gain arises not only from a strong backbone but also from iterative judge refinement. 

We selected Opus~5 because it was the strongest available backbone in our evaluation suite when the main experiments were conducted. GPT-6 \citep{openai2026gpt6astra} and Gemini~3.8 \citep{googledeepmind2026gemini38flash} were released near the completion of this work and were added subsequently; neither provides a clear improvement over the calibrated Opus~5 judge.

\vspace{0.15cm}
\textbf{Student Judge Model Structure.} \Cref{tab:student_judge_variants} compares eight student-judge variants under a controlled design. All variants share the same Qwen3-VL 2B backbone.

\textit{(i) PAI-AV pretraining} consistently improves judge-human alignment, increasing Pearson correlation by up to 8\% and reducing MAE by up to 16.7\%. This result indicates that embodiment-specific pretraining provides useful physical-world priors for judge learning.

\textit{(ii) Classifier-based judges} consistently outperform generative variants while requiring substantially less inference time. Within the generative paradigm, explicit reasoning provides only marginal gains over score with only \(+0.01\) $r$, increasing latency by more than \(5\times\). 

\textit{(iii) Predicting rubric subscores} achieves performance and latency comparable to directly predicting the overall score. However, the decomposed predictions provide finer-grained diagnostics and make the basis of each judgment more interpretable. We therefore adopt the PAI-AV-pretrained classifier with rubric-subscore outputs as the student judge configuration throughout our experiments.

\begin{figure}[t]
  \centering
  \includegraphics[width=\linewidth]{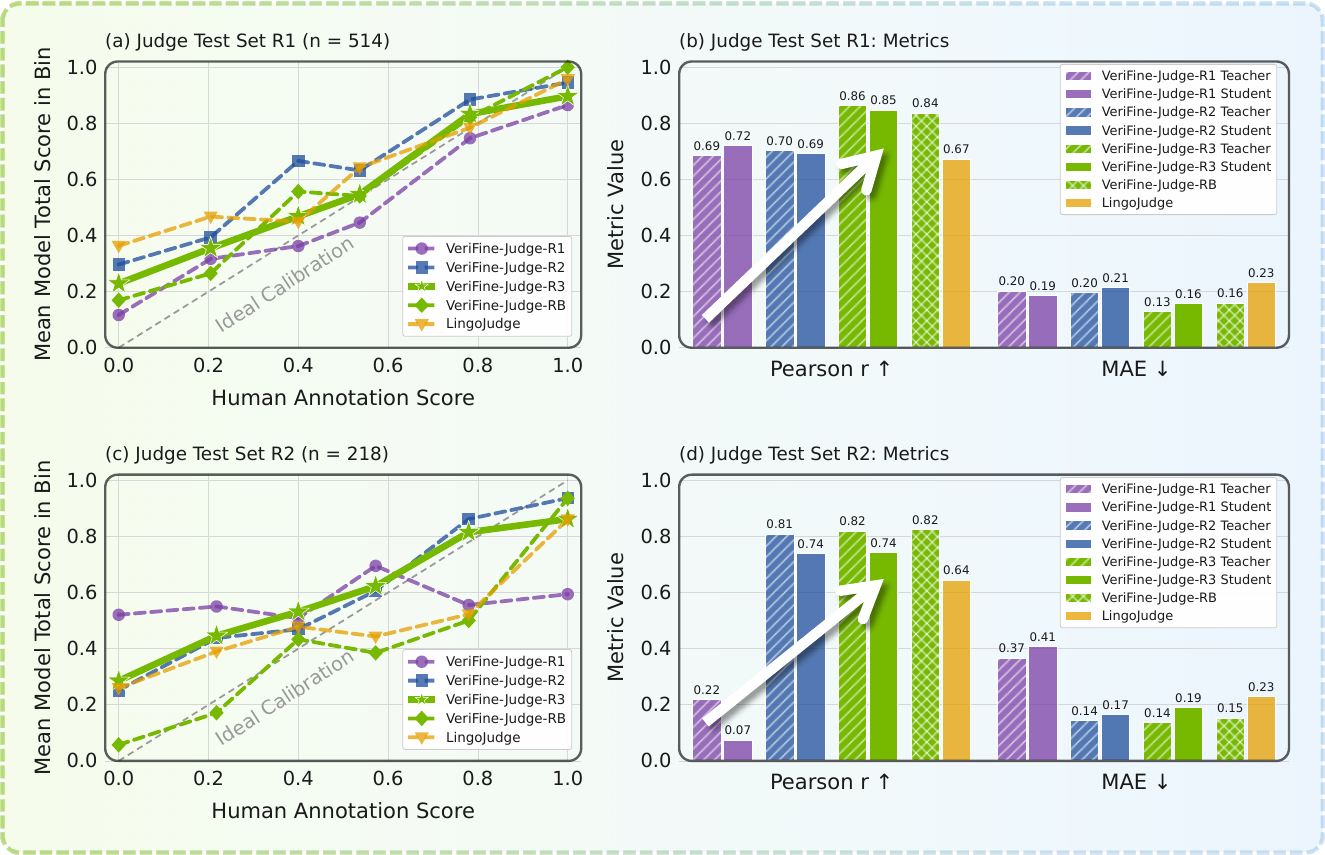}
  \vspace{0.1cm}
  \caption{\textbf{Evolved Judge Performance and Evolved Test Sets (R1 and R2).} \textcolor[HTML]{275317}{\textbf{(a) (c) Calibration curves}} showing the mean judge score within each human-score bin for two evolved test subsets for the judge test set; the gray diagonal denotes ideal calibration. \textcolor[HTML]{7030A0}{\textbf{(b) (d) Judge--human alignment}} measured by Pearson correlation and MAE for teacher and student judges across improvement rounds and reference-based baselines.}
  \label{fig:judge results of subsets}
\end{figure}

\vspace{0.15cm}
\textbf{Evolving Judge.}
\Cref{fig:judge results of subsets} reports judge performance on test subsets constructed across judge improvement rounds. Unlike a fixed benchmark, the cumulative judge evaluation set evolves with the policy: each round emphasizes newly exposed failure patterns and thereby reflects the current optimization target of \VeriFine. 

\textit{(i) Round 1}. It focuses primarily on nominal driving scenarios, which constitute the majority of the source distribution, and constructs judge evaluation and test subset R1. The resulting R1 student judge achieves a Pearson correlation of \(0.72\) on this subset. 

\textit{(ii) Round 2}. The failure analysis shifts the focus toward challenging scenarios, such as the pedestrians and obstacles encroaching into the ego lane. These policy-rollout cases form subset R2 and expose a substantial distribution shift: the R1 student obtains only \(r=0.07\), whereas the R2 student improves the correlation to \(0.74\). 

\textit{(iii) Round 3}. It introduces no additional evaluation subset; instead, it balances performance across the cumulative set and further targets known failure cases. This refinement increases the pooled Pearson correlation from \(0.71\) to \(0.82\) for the student judge, with the final teacher reaching \(0.85\). These results show that the Judge Improvement Loop progressively adapts verification to the policy's evolving failure distribution.

\subsection{VeriFine-Judge: Reference-free Rubric Judge}
\label{app:judge}

\begin{figure}[t]
  \centering
  \includegraphics[width=0.75\linewidth]{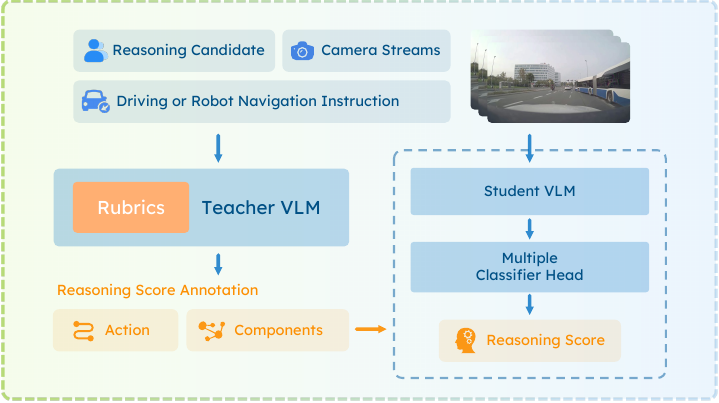}
  \caption{\textbf{Teacher--Student Architecture of the Reference-free Rubric Judge.} The teacher evaluates policy outputs using visual context, ego state, a driving or robot navigation instruction, and a structured rubric. Its evaluations are distilled into a compact student judge for scalable verification.}
  \label{fig:judge model}
  \vspace{0.3cm}
\end{figure}

Our reference-free judge evaluates candidate reasoning directly from the observable physical context, without requiring human annotation for reasoning during inference. Figure~\ref{fig:judge model} illustrates the teacher--student architecture. The teacher combines a frontier VLM with an explicit embodied-reasoning rubric. The rubric subscores and explanations support failure diagnosis and coactive calibration. Then, its structured evaluations are distilled into a compact VLM student for large-scale reward computation, curriculum construction, and test-time scaling.

\subsubsection{Teacher Judge Model}

We use Claude Opus 5 \citep{anthropic2026claudeopus5} as the teacher model. Its input contains three uniformly sampled history frames from the camera streams, ego-state information, the driving or robot navigation instructions, and the candidate policy output. The teacher returns multiple rubric scores (across two main dimensions: \textit{Action} and \textit{Component}), the coherence diagnostic, an overall score, and a concise explanation in a structured, constrained format.

\textbf{Driving Reasoning Rubric.}
For each sample, the driving reasoning judge predicts five primary binary rubric dimensions:

\begin{equation}
    \mathbf{z}_i^{\text{Driving}}
    =
    \left(
        z_i^{a_\mathrm{match}},
        z_i^{a_\mathrm{safe}},
        z_i^{c_\mathrm{ground}},
        z_i^{c_\mathrm{complete}},
        z_i^{c_\mathrm{causal}}
    \right).
    \label{eq:judge-dimensions}
\end{equation}

The first two \textit{Action} dimensions evaluate whether the proposed action is consistent with the driving instruction and safe under the observed scene. The remaining dimensions of \textit{Component} evaluate whether the reasoning grounds its claims in visible evidence, identifies all behavior-relevant scene components, and correctly explains their causal relationship to the proposed action. We additionally use reasoning coherence as a diagnostic dimension.

The teacher action score applies a hard gate:

\begin{equation}
    A_i^{\text{Driving}}
    =
    \begin{cases}
        0,
        & z_i^{a_\mathrm{match}} = 0, \\[2mm]
        \dfrac{1}{2}
        \left(
            z_i^{a_\mathrm{match}}
            +
            z_i^{a_\mathrm{safe}}
        \right),
        & \text{otherwise}.
    \end{cases}
    \label{eq:action-score}
\end{equation}

The critical-component score is computed as

\begin{equation}
    C_i^{\text{Driving}}
    =
    \frac{1}{3}
    \left(
        z_i^{c_\mathrm{ground}}
        +
        z_i^{c_\mathrm{complete}}
        +
        z_i^{c_\mathrm{causal}}
    \right).
    \label{eq:component-score}
\end{equation}

The overall score of the judge is

\begin{equation}
    s_i^{\text{Driving}}
    =
    0.6 A_i^{\text{Driving}} + 0.4 C_i^{\text{Driving}}.
    \label{eq:overall-judge-score}
\end{equation}

The action-match gate removes action credit when the proposed action is inconsistent with the driving instruction, while the separate component term preserves credit for correctly grounded scene understanding.

\textbf{Robot Navigation Reasoning Rubric.}
Navigation retains the \textit{Action} and \textit{Component} structure but evaluates goal-directed movement rather than action matching with driving instructions. The horizon of robot navigation instruction is longer than driving reasoning task, because it defines the final goal of the robot navigation. The judge predicts four binary rubric dimensions:

\begin{equation}
\mathbf{z}_i^{\mathrm{RoboNav}}
=
\left(
z_i^{a_{\mathrm{safe}}},
z_i^{a_{\mathrm{goal}}},
z_i^{c_{\mathrm{ground}}},
z_i^{c_{\mathrm{causal}}}
\right).
\label{eq:nav-judge-dimensions}
\end{equation}

Action safety assesses whether the proposed movement has sufficient clearance under the observed geometry without potential collision. Goal consistency assesses efficient progress toward the instructed destination, including purposeful exploration before target localization and appropriate stopping upon arrival. Naming the correct target does not compensate for an incorrect movement direction or a premature stop.

The \textit{Component} dimensions retain visual grounding and action--component causation, without a separate completeness score:

\begin{equation}
C_i^{\mathrm{RoboNav}}
=
\frac{1}{2}
\left(
z_i^{c_{\mathrm{ground}}}
+
z_i^{c_{\mathrm{causal}}}
\right).
\label{eq:nav-component-score}
\end{equation}

Unlike the driving action-match gate, navigation safety gates the entire score:

\begin{equation}
s_i^{\mathrm{RoboNav}}
=
z_i^{a_{\mathrm{safe}}}
\left(
0.6z_i^{a_{\mathrm{goal}}}
+
0.4C_i^{\mathrm{RoboNav}}
\right).
\label{eq:nav-overall-judge-score}
\end{equation}

Thus, unsafe movements receive zero overall credit, while safe movements are distinguished by goal progress and grounded justification. For instructions involving manipulation, the rubric evaluates navigation to the target rather than completion of the manipulation itself.

This task-specific aggregation reflects different reasoning demands. Driving decisions often hinge on a few critical scene elements, such as an emerging pedestrian or a red light. We therefore explicitly score component completeness and retain credit for correctly grounded and causally relevant observations even when the proposed action is unsafe, distinguishing preserved scene understanding from failures in action selection. In robot navigation, before the target is localized, selecting a productive exploration direction is more central than comprehensive component coverage. We therefore emphasize goal consistency, retain grounding and causation without a separate completeness term, and treat action safety as a prerequisite for overall credit.

\textbf{Coactive Calibration.} The initial rubric specifies the evaluation dimensions, their scoring criteria, common critical components, action priorities, and rules for handling noisy. 
When progress plateaus, human experts inspect the remaining disagreements together with the teacher's diagnostic scores and explanations. They may correct an individual dimension, identify missing physical evidence, or clarify an underspecified criterion. These clarifications are incorporated into the next rubric revision. Each revision is evaluated on both newly queried cases and cases retained from earlier iterations. We accept a revision only when it improves human alignment on the current batch without causing a substantial regression on previously calibrated cases.

\begin{table*}[t]
\centering
\caption{
\textbf{Comparison of Student Judge Variants.} All variants use Qwen3-VL as the backbone. We compare initialization with and without pretraining on our internal dataset, together with generation- and classifier-based output paradigms. Inference latency is averaged over 200 samples at batch size one on NVIDIA H100 GPUs, and the teacher model is accessed through a remote API.
}
\label{tab:student_judge_variants}
\small
\setlength{\tabcolsep}{13pt}
\renewcommand{\arraystretch}{1.5}

\begin{tabular}{c l l c c c}
\toprule
\multirow{2}{*}{\makecell{With PAI-AV\\Pretraining}} &
\multicolumn{2}{c}{Output Paradigm} &
\multirow{2}{*}{\makecell{Pearson $r\uparrow$}} &
\multirow{2}{*}{\makecell{MAE $\downarrow$}} &
\multirow{2}{*}{\makecell{Inference Latency\\(s/sample) $\downarrow$}} \\
\cmidrule(lr){2-3}
& Head & Output & & & \\
\midrule

\multirow{4}{*}{$\times$}
& \multirow{2}{*}{Generation}
& Reasoning $+$ Score & 0.72 & 0.18 & 2.34 \\
& & Score Only & 0.71 & 0.19 & 0.45 \\

\cmidrule(lr){2-6}
& \multirow{2}{*}{Classifier}
& Overall Score & 0.76 & 0.18 & 0.07 \\
& & Rubric Subscores & 0.76 & 0.19 & 0.07 \\

\midrule
\multirow{4}{*}{$\checkmark$}
& \multirow{2}{*}{Generation}
& Reasoning $+$ Score & 0.76 & \uline{0.16} & 2.40 \\
& & Score Only & 0.75 & 0.17 & 0.45 \\

\cmidrule(lr){2-6}
& \multirow{2}{*}{Classifier}
& Overall Score & \uline{0.81} & \textbf{0.15} & 0.08 \\
& & \cellcolor[HTML]{EDF4DB}Rubric Subscores
& \cellcolor[HTML]{EDF4DB}\textbf{0.82}
& \cellcolor[HTML]{EDF4DB}0.17
& \cellcolor[HTML]{EDF4DB}\textbf{0.07}%
\makebox[0pt][l]{%
  \hspace{0.35em}%
  {\scriptsize\textcolor{darkgreen}{$300\times$}}%
}
\\

\midrule
\rowcolor[HTML]{E8EDF3}
\multicolumn{3}{l}{Teacher Model - Opus 5 \citep{anthropic2026claudeopus5}}
& 0.85 & 0.13 & 20.92 \\
\bottomrule
\end{tabular}
\end{table*}

\subsubsection{Student Judge Model}

Although the teacher provides strong, structured evaluations, querying a frontier VLM throughout policy optimization is computationally expensive, with long latency and high API cost, especially for the reinforcement fine-tuning of the driving reasoning task based on GRPO with high query frequency. We therefore distill its evaluation capability into a compact student judge. The student receives the same physical context, driving instruction, and candidate output as the teacher and predicts all the rubric dimensions in a single forward pass.

The distillation dataset contains independent teacher-labeled samples. Each rubric dimension is modeled as a classification objective. The student training objective is equal-weight binary cross-entropy over all the rubric dimensions:

\begin{equation}
    \mathcal{L}_{\mathrm{judge}}
    =
    -\frac{1}{nB}
    \sum_{i=1}^{B}
    \sum_{k=1}^{n}
    \left[
        \widetilde{y}_{ik}\log p_{ik}
        +
        (1-\widetilde{y}_{ik})\log(1-p_{ik})
    \right],
\label{eq:student-loss}
\end{equation}

where $n$ is the number of the rubric dimensions, $p_{ik}$ is the predicted probability and $\widetilde{y}_{ik}$ is the teacher label, with $0.1$ label smoothing applied only to the action-match gate. The overall score is aggregated from the subscores and does not directly provide the supervision signal for training.

Using the student model of driving reasoning as an example, the student model processes samples approximately 300 times faster than the teacher model under the hardware settings described in \Cref{tab:student_judge_variants}, which demonstrates its real-time performance to support online training and evaluation at low cost. Moreover, we compared the different output heads for the student judge model and finally chose the classifier head with subscores to ensure interpretability with rubric scores, rather than the slower raw autoregressive generation paradigm.

\subsection{Evaluation and Testing Details}
\label{app:eval details}

Reliable evaluation is the foundation of scaling verification. This section describes how \VeriFine{} evaluates both the policy and the judge throughout self-improvement. We first introduce the policy and judge evaluation sets and then describe the evaluation protocols applied across iterations.

\begin{figure}[!h]
    \centering
    \vspace{0.2cm}
    \includegraphics[width=\linewidth]{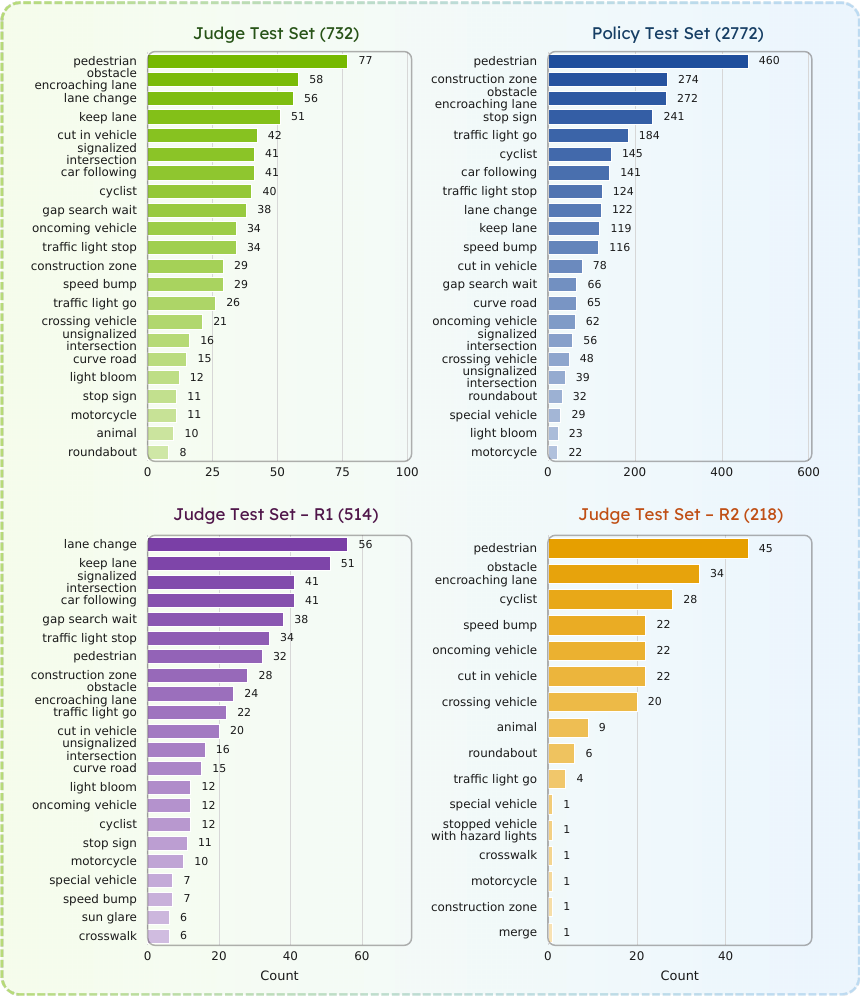}
    \caption{\textbf{Scenario Distribution of the Testing Sets in Driving Reasoning}. The figure shows the top 21 categories in each test set, which constitute the main body of the sets.}
    \label{fig:eval-set-distribution}
\end{figure}

\subsubsection{Driving Reasoning Evaluation and Testing}
\label{app:evaluation sets}

We construct two complementary, expert-verified evaluation and testing resources with distinct purposes. The fixed policy evaluation set $\mathcal{P}$ tracks policy progress and supports verification-boundary detection, whereas the cumulative judge evaluation set $\mathcal{J}_t$ supports human calibration and judge refinement. Because large-scale driving data inevitably contain noisy or erroneous samples, human experts inspect and verify the camera observations, trajectories, reasoning annotations, and reasoning scores of every evaluation sample. 

Notably, the final policy and judge performance are evaluated on the independent corresponding test sets $\mathcal{P}^\mathrm{test}$ and $\mathcal{J}_t^\mathrm{test}$ with the same distribution as the evaluation sets. The \Cref{tab:evaluation_sets} summarizes the details of those sets, and the \Cref{fig:eval-set-distribution} shows the distribution of the larger test set.

\begin{table}[h]
    \centering
    \small
    \caption{\textbf{Summary of the Evaluation and Testing Sets of Driving Reasoning}. The query batch $\mathcal{H}_t$ contains newly selected cases at judge-improvement iteration $t$, while $\mathcal{J}_t$ denotes the cumulative judge evaluation set.}
    \label{tab:evaluation_sets}
    \renewcommand{\arraystretch}{1.3} 
    \setlength{\tabcolsep}{12pt}
    \begin{tabularx}{\linewidth}{lllr}
        \toprule
        Set & Primary purpose & Construction & Size \\
        \midrule
        $\mathcal{P}$ \& $\mathcal{P}^\mathrm{test}$   & Policy monitoring and testing & Fixed, expert-checked        & 2134, 2772 \\
        $\mathcal{H}_1$ \& $\mathcal{H}_1^\mathrm{test}$  & Initial judge calibration and testing     & Broad base-policy rollouts   &   433, 514 \\
        $\mathcal{H}_2$ \& $\mathcal{H}_2^\mathrm{test}$ & Boundary-focused calibration and testing  & Updated-policy rollouts      &   179, 218 \\
        $\mathcal{J}_3$ \& $\mathcal{J}_3^\mathrm{test}$ & Final judge validation and testing        & $\mathcal{H}_1 \cup \mathcal{H}_2$,$\mathcal{H}_1^\mathrm{test} \cup \mathcal{H}_2^\mathrm{test}$ & 612, 732 \\
        \bottomrule
    \end{tabularx}
    \vspace{0.2cm}
\end{table}

\textbf{Policy Evaluation and Testing Set of Driving Reasoning}
\label{app:policy eval set}
The policy evaluation and testing sets $\mathcal{P}$ \& $\mathcal{P}^\mathrm{test}$ come from carefully selected driving scenarios. Human experts inspect each scenario and correct unreliable reasoning annotations. The resulting set covers diverse geographic regions, e.g., the U.S., Europe, the U.K., and South Korea, as well as varied road structures, weather conditions, and lighting conditions. It contains both reasoning-dense scenarios involving multiple interacting agents and reasoning-light scenarios with the straightforward decisions.

Because the reward judge evolves throughout self-improvement, its scores cannot provide a fixed basis for comparing policies across iterations. Human experts curate policy evaluation sets to maintain diversity across scenario categories and reasoning densities, providing a stable evaluation anchor for the complete self-improvement process. \Cref{fig:eval-set-distribution} summarizes the distribution of $\mathcal{P}^\mathrm{test}$ across geographic regions, scenario categories, and reasoning-density levels.

\vspace{0.1cm}
\textbf{Judge Evaluation and Testing Set of Driving Reasoning}
\label{app:judge eval set}
At judge-improvement iteration $t$, an incremental set $\mathcal{H}_t$ \& $\mathcal{H}_t^\mathrm{test}$ is constructed from outputs generated by the latest policy. As summarized in \Cref{tab:evaluation_sets}, the first batch contains broadly sampled cases generated by the initial policy and is used to establish the initial human--judge calibration. The second batch contains challenging cases generated after policy improvement. These cases emphasize newly exposed policy failures, judge--human disagreements, and outputs near the judge's effective verification boundary. Unlike policy training data, $\mathcal{J}_t$ deliberately retains both high- and low-quality outputs. Negative cases are necessary for identifying false positives, refining rubric boundaries, and detecting reward-hacking behavior during judge model training.

\subsubsection{Robot Navigation Reasoning Evaluation and Testing}
\label{app:reference based judge}

\vspace{0.2cm}
\textbf{Data Curation} 

VLNVerse \citep{lin2025vlnverse} provides interactive indoor environments and navigation episodes pairing language instructions with reference trajectories, including both fine-grained route-following and coarse-grained goal-directed tasks. These episode-level annotations do not directly provide the observation-aligned, decision-level reasoning supervision. We therefore collect robot-view observations and construct reasoning examples that connect a destination goal and local visual evidence to the next navigation decision. The target is to determine where to explore, how to approach the target, and when to stop, without receiving intermediate route instructions.

We collect egocentric image sequences along navigation trajectories in VLNVerse scenes. Given the original navigation instruction and sampled frames in the trajectory, Claude Opus~5 \citep{anthropic2026claudeopus5} segments each trajectory into consecutive turning, traversal, doorway, approach, and arrival intervals and generates a concise, landmark-grounded movement description for each interval. 

At each interval's starting point, we pair the generated annotation with the current observation, two historical observations, and a destination instruction derived from the original task by retaining the final goal and removing intermediate route commands. This produces 19k samples with resolved destination instructions. The resulting pool supports judge-guided selection of reasoning supervision for policy improvement.

\begin{wraptable}{r}{0.41\textwidth}
    \centering
    \footnotesize
    \vspace{-0.5cm}
    \caption{\textbf{Robot Navigation Evaluation and Test Sets.}}
    \vspace{-0.2cm}
    \label{tab:navigation_evaluation_sets}
    \renewcommand{\arraystretch}{1.2}
    \setlength{\tabcolsep}{3pt}
    \begin{tabularx}{\linewidth}{@{}Xrr@{}}
        \toprule
        Set & Evaluation & Test \\
        \midrule
        Policy ($\mathcal{P}$)             & 182 & 220 \\
        Initial judge ($\mathcal{H}_1$)    & 175 & 211 \\
        R2 added judge cases ($\mathcal{H}_2$)      & 37  & 70  \\
        Cumulative judge ($\mathcal{J}_2$) & 212 & 281 \\
        \bottomrule
    \end{tabularx}
\end{wraptable}

\textbf{Policy Evaluation Set of Robot Navigation Reasoning}

The fixed policy evaluation set and independent policy test set contain 182 and 220 samples, respectively, from 46 episodes across 28 scenes. The distribution of the larger policy test set is 53 turning, 69 traversal, 36 doorway, 36 approach, and 26 arrival decisions. All the samples in the policy set have been verified to ensure the quality of the decision time and reasoning content.

\textbf{Judge Evaluation Set of Robot Navigation Reasoning}
We initially construct a candidate pool by pairing source reasoning annotations with VLM-generated negative candidates. These candidates introduce controlled errors in the proposed movement or referenced scene elements while preserving the instruction and observations. Human review determines their rubric scores and yields an initial judge evaluation and independent test sets of 175 and 211 candidates.

Subsequent policy rollouts reveal weaknesses in exploration efficiency, particularly in selecting productive directions and deciding whether to approach, stop, or continue searching. Guided by these observed failure patterns, the agent constructs 107 additional contrast candidates, targeting distinctions between superficially plausible actions and movements that actually advance the navigation goal. The expanded judge evaluation and test set contain 212 and 281 candidates.

\subsubsection{VeriFine-Judge-RB: Reference-Based Rubric Judge}
\label{app:reference based judge}

Because the reference-free judge evolves across iterations, we use a fixed reference-based judge $J^{\mathrm{ref}}$ to test the policy progress on $\mathcal{P}^\mathrm{test}$, which compares each policy-generated reasoning with its expert-verified reference reasoning. VeriFine-Judge-RB remains fixed during development monitoring and final policy evaluation. Its use as a reward model is confined to the reference-based baselines explicitly identified in \Cref{tab:policy_improvement}. The evolving reference-free judge supplies the training reward for VeriFine policies.

The VeriFine-Judge-RB is a DeBERTa-v3-base cross-encoder \citep{he2021debertav3} initialized from Lingo-Judge \citep{marcu2024lingoqa}. It takes a fixed question, the reference reasoning, and the candidate reasoning as input, and produces a scalar score in $[0,1]$ through a single classification head. VeriFine-Judge-RB is further fine-tuned on approximately 500K reference--candidate pairs using binary cross-entropy. 

We also leverage the rubric across action and critical components for the teacher model to obtain large-scale reasoning annotations to train the VeriFine-Judge-RB model. We leverage Claude Opus, GPT, and Gemini to annotate the reasoning score individually based on the rubric, and average them as the training supervision data for the VeriFine-Judge-RB model.

\subsection{Experiment Details}
\label{app:exp}

\subsubsection{Evaluation Metrics}
\label{app:evaluation_metrics}

\vspace{0.1cm}
\textbf{Reasoning Score.}
For each sample, the evaluation protocol assigns a reasoning score $s_i\in[0,1]$: the reference-based judge uses expert reference reasoning, the reference-free judge uses the observed context and rubric. We report the mean scaled by 100.
\begin{equation}
\mathrm{Reasoning\ Score}
=
\frac{100}{N}\sum_{i=1}^{N}s_i.
\end{equation}

\textbf{Judge--Human Alignment.}
Judge capability is evaluated by comparing the predicted overall reasoning scores \(s_i\) with expert scores \(h_i\). We report Pearson correlation $r$, which measures agreement in relative sample quality, and mean absolute error (MAE),
\begin{equation}
\mathrm{MAE}
=
\frac{1}{N}\sum_{i=1}^{N}|s_i-h_i|,
\end{equation}
which measures absolute calibration error. Higher Pearson correlation $r$ and lower MAE indicate better alignment with human evaluation.

\textbf{Trajectory Accuracy.}
For the $k$-th predicted trajectory, average displacement error is
\begin{equation}
\mathrm{ADE}^{(k)}
=
\frac{1}{T}\sum_{t=1}^{T}
\left\|
\hat{\mathbf{p}}_{t}^{(k)}-\mathbf{p}_{t}
\right\|_2,
\end{equation}
where $\hat{\mathbf{p}}_{t}^{(k)}$ and $\mathbf{p}_{t}$ are the predicted and ground-truth positions at timestep $t$, respectively. We report ADE for a single predicted trajectory and
\begin{equation}
\mathrm{minADE}_6
=
\min_{k\in\{1,\ldots,6\}}
\mathrm{ADE}^{(k)}.
\end{equation}
Both metrics are averaged over evaluation samples and measured in meters; lower values indicate better trajectory accuracy.

\subsubsection{Implementation Details}
\label{app:implementation}

\vspace{0.2cm}
\textbf{Reinforcement Fine-tuning for Driving Policy.}
In the driving reasoning task, we initialize the Alpamayo~1.5 8B policy and optimize it with reinforcement fine-tuning with GRPO, using VeriFine-Judge both to provide the online reward and to adaptively construct the training curriculum. The selected training set with curriculum judge contains the fixed 25K samples from a pool of 400K candidates, and the training data budget is fixed for fair comparison. Each update uses 32 prompts with \(K=16\) rollouts per prompt, resulting in an effective batch size of 512 rollouts. We train for fixed 2,700 steps using AdamW with a learning rate of \(1\times10^{-5}\) and a KL coefficient of \(0.02\), on 20 NVIDIA H100 GPU nodes. 

\textbf{Supervised Fine-tuning for Robot Navigation Policy.}
For the robot navigation reasoning task, the policy is initialized from the Qwen3-VL 2B base model. All runs use 8 NVIDIA H100 GPUs and a global batch size of 128. We freeze the vision backbone while updating the language model and visual merger with AdamW, using learning rates of $10^{-5}$ and $10^{-6}$, respectively, and weight decay of $0.01$. Moreover, the SFT uses cosine learning-rate decay.

\textbf{Student VeriFine-Judge Training.}
We freeze its vision backbone and train the language backbone and a five-way classification head on 35K teacher-labeled examples using equal-weight binary cross-entropy, with \(0.1\) label smoothing applied only to the action-matching output. We train it on 8 NVIDIA H100 GPUs with a global batch size of 16, using AdamW with learning rates of \(4\times10^{-5}\) and \(10^{-3}\) for the backbone and classification head, respectively. We use 500 warmup steps followed by cosine decay to \(0.2\times\) the peak learning rate, weight decay of \(0.1\), and gradient clipping at \(1.0\).

\textbf{VeriFine-Judge-RB Training.}
We train our reference-based model on approximately 500K reference--candidate reasoning pairs with 16 NVIDIA H100 GPUs, using binary cross-entropy with rubric-derived soft targets and a maximum sequence length of 256. We use fused AdamW with a global batch size of 128, a learning rate of \(4\times10^{-5}\), weight decay \(0.01\), 250 warmup steps, cosine decay to \(1\%\) of the peak learning rate, and gradient clipping at \(1.0\).

\subsubsection{Loop Execution Protocol}
\label{app:loop_protocol}

\vspace{0.2cm}
\textbf{Judge Improvement Loop Trigger.}
Boundary checks are performed every 150 steps on the policy evaluation set $\mathcal{P}$ during policy training. $P_t$ is measured by the fixed VeriFine-Judge-RB evaluator. The progress signals use the reasoning score, with $\Delta R_t^{\mathrm{judge}}$ computed using the current judge, which is used as the reward judge. We use a window of $K=4$ and a 0.5 threshold.

\textbf{Curriculum Selection.}
The executed selection rule is a ranking rule based on target-quality, learnability, and category type, with a tie-breaking rule, yielding 25K examples from the 400K candidate pool for driving. For navigation, context priority is determined by a failure-based selection rule, while demonstration quality is checked using the target-quality criterion. We retain 13K examples per round. Selection rules are updated from failures observed on $\mathcal{P}$.

\textbf{Rubric Revision.}
The agent modifies sub-rubric details and demonstration examples while keeping the rubric dimensions unchanged. Each proposal is evaluated on the current calibration batch and retained calibration cases. Revisions are accepted according to Pearson correlation and MAE performance. Each round permits 20 version rubrics, and human review is requested when the average improvement in Pearson correlation is less than 0.2 for more than ten iterations.

\textbf{Human Calibration and Test Annotation}
During coactive calibration, human annotators and the agent resolve rubric ambiguities through evidence-based discussion, allowing both the rubric and human calibration scores to be revised. Then, we trained another set of annotators and independently scored the judge test set without access to judge scores. Test labels are subsequently frozen and used consistently across all iteration rounds and judge comparisons, without informing rubric refinement or model selection.

\subsection{Limitations and Future Work}

While the judge evaluation set evolves with the policy to capture newly exposed failure patterns, the current framework retains a fixed policy evaluation set as a consistent anchor for measuring improvement across iterations. Future work could preserve a fixed held-out test set for comparable reporting while introducing an adaptive policy evaluation set that co-evolves with the policy and judge to provide increasingly informative signals for failure discovery and curriculum construction. The framework also still relies on reasoning-annotated data: although the reference-free VeriFine-Judge can filter low-quality reasoning during training, the reference reasoning used for policy evaluation remains human-verified. A natural extension is therefore to jointly evolve a reasoning annotator that generates and refines reference reasoning using structured judge feedback. Finally, while coactive calibration already consolidates human expertise into the evolving rubric and reduces repeated queries, explicitly distilling human corrections, instructions, and rubric revisions into a reusable guidance model could further reduce human involvement and eventually enable continued judge improvement without routine human guidance.

\end{document}